\pdfoutput=1
\documentclass[letterpaper, 10 pt, journal, twoside]{IEEEtran}

\IEEEoverridecommandlockouts                              %

\usepackage[backend=biber,
            url=false,
            isbn=false,
            doi=false,
            backref=false,
            style=ieee,
            natbib=true,%
            mincitenames=1,
            maxcitenames=1,
            citestyle=numeric-comp,
            sorting=nyt,
            block=none]{biblatex}
        
\renewcommand{\bibfont}{\small}
\usepackage{graphics}
\usepackage{dsfont}
\usepackage[pdftex]{graphicx}
\usepackage{wrapfig}
\DeclareGraphicsExtensions{.pdf,.png,.jpg}
\usepackage{epsfig}
\usepackage[font={small}]{caption}
\usepackage{subcaption}
\usepackage[rightcaption]{sidecap}
\usepackage{pbox}
\usepackage{cuted}

\usepackage{bigstrut}
\usepackage{mathtools}
\usepackage{amsmath, amssymb, amscd}
\usepackage{ wasysym } %
\usepackage{mathptmx} %
\usepackage{gensymb} 
\usepackage{nicefrac}       %
\numberwithin{equation}{section} 

\DeclareMathAlphabet{\mathcal}{OMS}{lmsy}{m}{n}
\usepackage{textcomp} %

\usepackage{algorithm} %
\usepackage{algorithmicx, algpseudocode} %

\usepackage{array} %
\usepackage{tabularx}
\usepackage{multirow}
\usepackage{multicol}
\usepackage{booktabs}
\usepackage{tabulary}

\usepackage[utf8]{inputenc}
\usepackage[english]{babel} %
\usepackage{units}
\usepackage{bm}
\usepackage{xspace}
\usepackage{flushend}%
\usepackage{balance} 
\usepackage{csquotes}
\usepackage{makeidx}
\usepackage{blindtext}

\usepackage{enumitem}

\usepackage{ragged2e}
\usepackage{soul} %
\usepackage{subfiles} %

\usepackage{url}
\makeatletter
\g@addto@macro{\UrlBreaks}{\UrlOrds}
\makeatother
\usepackage{color}
\usepackage[usenames,dvipsnames,table,xcdraw]{xcolor}
\usepackage{pgfplots} %
\pgfplotsset{compat=newest}

\usepackage{marginnote}
\usepackage{soul} %

\usepackage[colorinlistoftodos]{todonotes}
\newcommand{\tocite}[1]{%
\textcolor{red}{[cite:\ifthenelse{\equal{#1}{}}{}{#1}?]}
}

\newcommand{\ignore}[1]{}

\renewcommand{\baselinestretch}{1.0}


\usepackage{xspace}
\usepackage[normalem]{ulem}

\newcommand{\algname}{Recovery RL\xspace}

\newcommand{\qsafe}{Q_{\mathrm{risk}}^\pi}
\newcommand{\vsafe}{V_{\mathrm{risk}}^\pi}
\newcommand{\qsafehat}{\hat{Q}_{\phi, \mathrm{risk}}^\pi}

\newcommand{\jsafe}{J_{\mathrm{risk}}(s_t, a_t, s_{t+1}; \phi)}
\newcommand{\gsafe}{\gamma_{\mathrm{risk}}}
\newcommand{\esafe}{\epsilon_{\mathrm{risk}}}
\newcommand{\safeset}{\mathcal{T}^\pi_{\mathrm{safe}}}
\newcommand{\recset}{\mathcal{T}^\pi_{\mathrm{rec}}}

\newcommand{\pirec}{\pi_{\mathrm{rec}}}
\newcommand{\pitask}{\pi_{\mathrm{task}}}
\newcommand{\demos}{\mathcal{D}_{\mathrm{offline}}}

\DeclareMathOperator*{\argmax}{arg\,max}

\title{
Recovery RL: Safe Reinforcement Learning \\with Learned Recovery Zones
}

\author{
Brijen Thananjeyan$^{*1}$, Ashwin Balakrishna$^{*1}$, Suraj Nair$^{2}$, Michael Luo$^{1}$, Krishnan Srinivasan$^{2}$,\\Minho Hwang$^{1}$,
Joseph E. Gonzalez$^{1}$, Julian Ibarz$^{3}$, Chelsea Finn$^{2}$, Ken Goldberg$^{1}$
\thanks{* equal contribution}
\thanks{Manuscript received: October, 15, 2019; Revised February 2, 2020; Accepted February 27, 2020.}%
\thanks{This paper was recommended for publication by Editor Dana Kulic upon evaluation of the Associate Editor and Reviewers' comments.}
\thanks{$^{1}$Brijen Thananjeyan, Ashwin Balakrishna, Suraj Nair, Michael Luo, Minho Hwang, Joseph E. Gonzalez, and Ken Goldberg are with the Dept. of Electrical Engineering and Computer Science, University of California, Berkeley, USA
        {\tt\footnotesize \{bthananjeyan, ashwin\_balakrishna, michael.luo, gkgkgk1215, jegonzal, goldberg\}@berkeley.edu}}%
\thanks{$^{2} $Suraj Nair, Krishnan Srinivasan, and Chelsea Finn are with the Dept. of Computer Science, Stanford University, USA
        {\tt\footnotesize \{surajn, krshna, cbfinn\}@stanford.edu}}
\thanks{$^{3} $Julian Ibarz is with Google AI, USA
        {\tt\footnotesize julianibarz@google.com}}
\thanks{Digital Object Identifier (DOI): see top of this page.}

}

\begin{document}

\maketitle

\begin{abstract}
Safety remains a central obstacle preventing widespread use of RL in the real world: learning new tasks in uncertain environments requires extensive exploration, but safety requires limiting exploration. We propose \algname, an algorithm which navigates this tradeoff by (1) leveraging offline data to learn about constraint violating zones \emph{before} policy learning and (2) \emph{separating} the goals of improving task performance and constraint satisfaction across two policies: a task policy that only optimizes the task reward and a recovery policy that guides the agent to safety when constraint violation is likely. We evaluate \algname on 6 simulation domains, including two contact-rich manipulation tasks and an image-based navigation task, and an image-based obstacle avoidance task on a physical robot. We compare \algname to 5 prior safe RL methods which jointly optimize for task performance and safety via constrained optimization or reward shaping and find that \algname outperforms the next best prior method across all domains. Results suggest that \algname trades off constraint violations and task successes 2 - 20 times more efficiently in simulation domains and 3 times more efficiently in physical experiments. See \url{https://tinyurl.com/rl-recovery} for videos and supplementary material.

\end{abstract}
\begin{IEEEkeywords}
Reinforcement Learning, Safety%
\end{IEEEkeywords}
\section{Introduction}
\label{sec:intro}
Reinforcement learning (RL) provides a general framework for robots to acquire new skills, and has shown promise in a variety of robotic domains such as navigation~\cite{safe-visual-navigation},
locomotion~\cite{SAC-applications}, and manipulation~\cite{qt-opt, dexterous-manip}. However, when deploying RL agents in the real world, unconstrained exploration can result in highly suboptimal behaviors which can damage the robot, break surroundings objects, or bottleneck the learning process. For example, consider an agent tasked with learning to extract a carton of milk from a fridge. If it tips over the carton, then not only can this possibly break the carton and create a mess, but it also requires laborious human effort to wipe up the milk and replace the carton so that the robot can continue learning. In the meantime, the robot is not able to collect experience or improve its policy until the consequences of this violation are rectified. Thus, endowing RL agents with the ability to satisfy constraints during learning not only enables robots to interact safely, but also allows them to more efficiently learn in the real world. However, enforcing constraints on the agent's behavior during learning is challenging, since system dynamics and the states leading to constraint violations may be initially unknown and must be learned from experience, especially when learning from high dimensional observations such as images. Safe exploration poses a tradeoff: learning new skills through environmental interaction requires exploring a wide range of possible behaviors, but learning safely forces the agent to restrict exploration to constraint satisfying states.

\begin{figure}
     \centering
     \includegraphics[width=.47\textwidth]{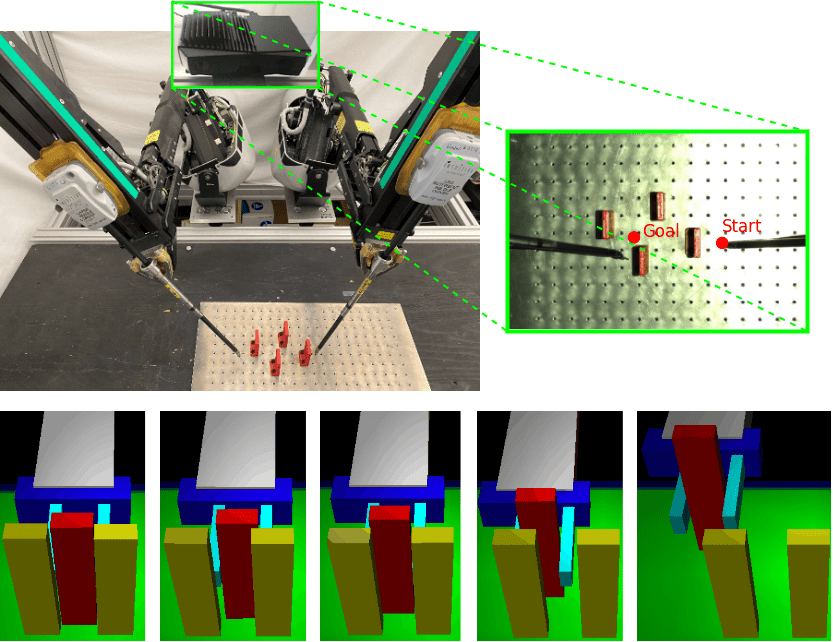}
     \caption{\algname can safely learn policies for contact-rich tasks from high-dimensional image observations in simulation experiments and on a physical robotic system. We evaluate \algname  on an image-based obstacle avoidance task with delta-position control on the da Vinci Research Kit (top left) with overhead image observations (top right). We find that \algname substantially outperforms prior methods (Figure~\ref{fig:reacher}), suggesting that it can be used for visuomotor control on physical robots. We also find that \algname can perform challenging contact-rich manipulation tasks in simulation; as shown in the bottom row, \algname successfully extracts the red block without toppling other blocks by learning to nudge it away from other blocks before grasping it.}
        \label{fig:robot}
\end{figure}

We consider a RL formulation subject to constraints on the probability of unsafe future behavior and design an algorithm that can balance the often conflicting objectives of task-directed exploration and safety. Most prior work in safe RL integrates constraint satisfaction into the task objective to jointly optimize the two. While these approaches are appealing for their generality and simplicity, there are two key aspects which make them difficult to use in practice. First, the inherent objective conflict between exploring to learn new tasks and limiting exploration to avoid constraint violations can lead to suboptimalities in policy optimization. Second, exploring the environment to learn about constraints requires a significant amount of constraint violations during learning. However, this can result in the agent taking uncontrolled actions which can damage itself and the environment.

We take a step towards addressing these issues with two key algorithmic ideas. First, inspired by recent work in robust control~\cite{safety-framework, HJ-reachability, shielding, MPC-shielding}, we represent the RL agent with two policies: the first policy focuses on optimizing the unconstrained task objective (task policy) and the second policy takes control when the task policy is in danger of constraint violations in the near future (recovery policy). Instead of modifying the policy optimization procedure to encourage constraint satisfaction, which can introduce suboptimality in the learned task policy~\cite{safety-gym}, the recovery policy can be viewed as defining an alternate MDP for the task policy to explore in which constraint violations are unlikely. Separating the task and recovery policies makes it easier to balance task performance and safety, and allows using off-the-shelf RL algorithms for both. 
Second, we leverage offline data to learn a recovery set, which indicates regions of the MDP in which future constraint violations are likely, and a recovery policy, which is queried within this set to prevent violations. This offline data can be collected under human supervision to illustrate examples of desired behaviors before the agent interacts with the environment or can contain unsafe behaviors previously experienced by the robot in the environment when performing other tasks. Both the recovery set and policy are updated online with agent experience, but the offline data allows the agent to observe constraint violations and learn from them without the task policy directly having to experience too many uncontrolled violations during learning.

We present \algname, a new algorithm for safe robotic RL. Unlike prior work, \algname (1) can leverage offline data of constraint violations to learn about constraints \emph{before} interacting with the environment, and (2) uses separate policies for the task and recovery to learn safely without significantly sacrificing task performance. We evaluate \algname against 5 state-of-the-art safe RL algorithms on 6 navigation and manipulation domains in simulation, including a visual navigation task, and find that \algname trades off constraint violations and task successes 2 - 20 times more efficiently than the next best prior method. We evaluate \algname on an image-based obstacle avoidance task on a physical robot and find that it trades off constraint violations and task successes 3 times more efficiently than the next best prior algorithm.

\begin{figure*}
     \centering
         \includegraphics[width=\textwidth]{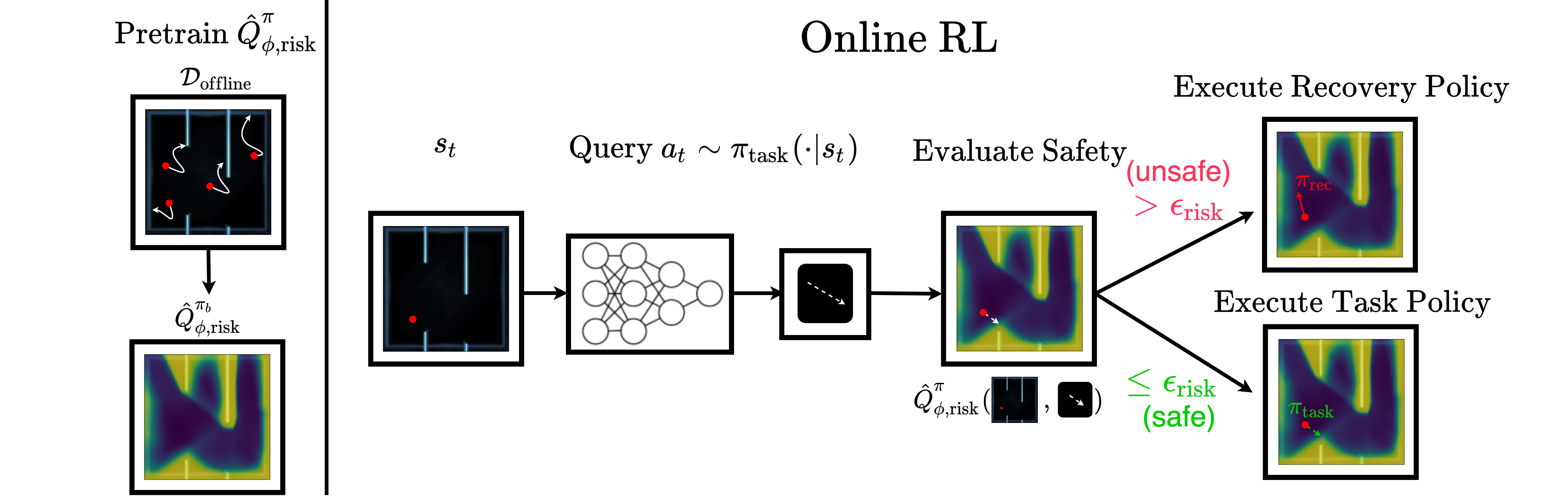}
        \caption{\textbf{\algname: }For intuition, we illustrate \algname on a 2D maze navigation task where a constraint violation corresponds to hitting a wall. \algname first learns safety critic $\qsafehat$ with offline data from some behavioral policy $\pi_b$, which provides a small number of controlled demonstrations of constraint violating behavior as shown on the left. For the purposes of illustration, we visualize the average of the $\qsafehat$ learned by \algname over 100 action samples. Then, at each timestep, \algname queries the task policy $\pitask$ for some action $a$ at state $s$, evaluates $\qsafehat(s, a)$, and executes the recovery policy $\pirec$ if $\qsafehat(s, a) > \esafe$ and $\pitask$ otherwise. The task policy, recovery policy, and safety critic are updated after each transition from agent experience.}
        \label{fig:splash}
\end{figure*}

\section{Related Work}
\label{sec:RW}
Prior work has studied safety in RL in several ways, including imposing constraints on expected return~\cite{CPO, RCPO}, risk measures~\cite{risk-RL, risk-sensitive-RL, pg-conditional-values, WCPG}, and avoiding regions of the MDP where constraint violations are likely~\cite{leave-no-trace, bridging-safety-RL, HJ-reachability, SAVED, StabilitySafeMBRL, SafeExpGP}. We build on the latter approach and design an algorithm which uses a learned recovery policy to keep the RL agent within a learned safe region of the MDP.

\textbf{Jointly Optimizing for Task Performance and Safety: }A popular strategy in algorithms for safe RL involves modifying the policy optimization procedure of standard RL algorithms to simultaneously reason about both task reward and constraints using methods such as trust regions~\cite{CPO}, optimizing a Lagrangian relaxation~\cite{RCPO, risk-sensitive-control,learn-to-be-safe}, or constructing Lyapunov functions~\cite{lyapunov-safety1, lyapunov-safety2}. The most similar of these works to \algname is~\citet{learn-to-be-safe}, which trains a safety critic to estimate the probability of future constraint violation under the current policy and optimizes a Lagrangian objective function to limit the probability of constraint violations while maximizing task reward. Unlike~\citet{learn-to-be-safe}, which uses the safety critic to modify the task policy optimization objective, \algname uses it to determine when to execute a learned recovery policy which minimizes the safety critic to keep the agent in safe regions of the MDP. This idea enables \algname to more effectively balance task performance and constraint satisfaction than algorithms which jointly optimize for task performance and safety.

\textbf{Restricting Exploration with an Auxiliary Policy: }Another approach to safe RL explicitly restricts policy exploration to a safe subset of the MDP using a recovery or shielding mechanism. This idea has been explored in \cite{safety-framework, HJ-reachability}, which utilize Hamilton-Jacobi reachability analysis to define a task policy and safety controller, and in the context of shielding~\cite{shielding, MPC-shielding, shielding-formal-logic}.
In contrast to these works, which assume approximate knowledge of system dynamics or require precise knowledge of constraints apriori, \algname learns information about the MDP, such as constraints and dynamics, from a combination of offline data and online experience. This allows \algname to scale to high-dimensional state spaces such as images, in which exact specification of system dynamics and constraints can be very challenging, and is often impossible. Additionally, \algname reasons about chance constraints rather than robust constraints, which may be challenging to satisfy when dynamics are unknown.
~\citet{safety-framework} design and prove safety guarantees for learning-based controllers in a robust optimal control setting with known dynamics and a robust control invariant safe set. With these additional assumptions, \algname has similar theoretical properties as well.~\citet{compound-RL} and~\citet{leave-no-trace} introduce reset policies which are trained jointly with the task policy to reset the agent to its initial state distribution, ensuring that the task policy only learns behaviors which can be reset~\cite{leave-no-trace}. However, enforcing the ability to fully reset can be impractical or inefficient. Inspired by this work, \algname instead executes approximate resets to nearby safe states when constraint violation is probable.~\citet{safe-visual-navigation} learns the probability of constraint violation conditioned on an action plan to activate a hand-designed safety controller. In contrast, \algname uses a learned recovery mechanism which can be broadly applied across different tasks.

\textbf{Leveraging Demonstrations for Safe RL and Control: }There has also been significant prior work investigating how demonstrations can be leveraged to enable safe exploration. \citet{LearningMPC, abc-lmpc} introduce model predictive control algorithms which leverage initial constraint satisfying demonstrations to iteratively improve their performance with safety guarantees and \citet{SAVED} extends these ideas to the RL setting. In contrast to these works, \algname learns a larger safe set that explicitly models future constraint satisfaction and also learns the problem constraints from prior experience without task specific demonstrations. Also, \algname is compatible with model-free RL algorithms while \cite{SAVED, abc-lmpc} require a dynamics model to evaluate reachability-based safety online.
\section{Problem Statement}
\label{sec:ps}
We consider RL under Markov decision processes (MDPs), which can be described by tuple $\mathcal{M} = (\mathcal{S}, \mathcal{A}, P(\cdot|\cdot,\cdot), R(\cdot, \cdot), \gamma, \mu)$ where $\mathcal{S}$ and $\mathcal{A}$ are the state and action spaces. Stochastic dynamics model $P: \mathcal{S} \times \mathcal{A} \times \mathcal{S} \rightarrow [0, 1]$ maps a state and action to a probability distribution over subsequent states, $\gamma \in [0, 1]$ is a discount factor, $\mu$ is the initial state distribution ($s_0\sim\mu$), and $R: \mathcal{S} \times \mathcal{A} \rightarrow \mathbb{R}$ is the reward function. We augment the MDP with an extra constraint cost function $C: \mathcal{S} \rightarrow \{0, 1\}$ which indicates whether a state is constraint violating and associated discount factor $\gsafe \in [0, 1]$. This yields the following new MDP: $(\mathcal{S}, \mathcal{A}, P(\cdot|\cdot,\cdot), R(\cdot, \cdot), \gamma, C(\cdot), \gsafe)$. We assume that episodes terminate on violations, equivalent to transitioning to a constraint-satisfying absorbing state with zero reward.

Let $\Pi$ be the set of Markovian stationary policies. Given policy $\pi \in \Pi$, the expected return is defined as $R^\pi = \mathbb{E}_{\pi,\mu, P}\left[ \sum_t \gamma^t R(s_t, a_t)\right]$ 
and the expected discounted probability of constraint violation is defined as  $\qsafe(s_i, a_i) = \mathbb{E}_{\pi,\mu, P}\left[ \sum_t \gsafe^t C(s_{t+i})\right] = \sum_t \gsafe^t\mathbb{P}\left(C(s_{t+i}) = 1\right)$, which we would like to be below a threshold $\esafe \in [0, 1]$.  The goal is to solve the following constrained optimization problem:
\begin{align}
\label{eq:objective}
    \pi^* = \argmax_{\pi \in \Pi} \{R^\pi : \qsafe(s_0, a_0) \leq \esafe\}
\end{align}
This setting exactly corresponds to the CMDP formulation from~\cite{def:cmdp}, but with constraint costs limited to binary indicator functions for constraint violating states. We limit the choice to binary indicator functions, as they are easier to provide than shaped costs and use $\qsafe$ to convey information about delayed constraint costs. We define the set of feasible policies, $\left\{\pi:\qsafe\leq \epsilon\right\}$, the set of $\epsilon$-safe policies $\Pi_{\epsilon}$. Observe that if $\gsafe=1$, then by the assumption of termination on constraint violation, $\qsafe(s_i, a_i) = \mathbb{P}\left(\bigcup_t C(s_t) = 1\right)$, or the probability of a constraint violation in the future. Setting $\esafe=0$ as well results in a robust optimal control problem.

We present an algorithm to optimize equation~\eqref{eq:objective} by utilizing a pair of policies, a \textit{task policy} $\pitask$, which is trained to maximize $R^\pi$ over ${\pitask \in \Pi}$ and a \textit{recovery policy} $\pirec$, which attempts to guide the agent back to a state-action tuple $(s, a)$ where $\qsafe (s, a) \leq \esafe$. We assume access to a set of transitions from offline data ($\demos$) with examples of constraint violations. Unlike in typical imitation learning settings, this data need not illustrate task successes, but shows possible ways to violate constraints. We leverage $\demos$ to constrain exploration of the task policy to reduce the probability of constraint violation during environment interaction.
\vspace{-0.1in}
\section{Recovery RL}
\label{sec:alg}
We outline the central ideas behind \algname. In Section~\ref{subsec:safety-critic}, we review how to learn a safety critic to estimate the probability of future constraint violations for the agent's policy. Then in Section~\ref{subsec:recovery-region}, we show how this safety critic is used to define the recovery policy for \algname and the recovery set in which it is activated. In Section~\ref{subsec:offline-pretraining} we discuss how the safety critic and recovery policy are initialized from offline data and in Section~\ref{subsec:implementation} we discuss implementation details. See Algorithm~\ref{alg:main} and Figure~\ref{fig:splash} for further illustration of \algname.
\vspace{-0.15in}
\subsection{Preliminaries: Training a Safety Critic}
\label{subsec:safety-critic}
As in~\citet{learn-to-be-safe}, \algname learns a critic function $\qsafe$ that estimates the discounted future probability of constraint violation of the current policy $\pi$:
\begin{align}
    \label{eq:q-safe-def}
    \begin{split}
    &\qsafe(s_t, a_t) = \mathbb{E}_\pi\left[\sum_{t'=t}^\infty \gsafe^{t'-t} c_{t'}|s_t, a_t\right]\\
    &= c_t + (1 - c_t)\gsafe\mathbb{E}_\pi\left[\qsafe(s_{t+1}, a_{t+1})|s_t, a_t\right].
    \end{split}
\end{align}
Here $c_t=1$ indicates that state $s_t$ is constraint violating with $c_t=0$ otherwise. Note we do not assume access to the true constraint cost function $C$. This is different from the standard Bellman equations to the assumption that episodes terminate when $c_t=1$. In practice, we train a sample-based approximation $\qsafehat$, parameterized by $\phi$, by approximating these equations using sampled transitions $(s_t, a_t, s_{t+1}, c_t)$.

We train $\qsafehat$ by minimizing the following MSE loss with respect to the target (RHS of equation~\ref{eq:q-safe-def}).
\begin{align}
    \label{eq:safety_critic_loss}
    \begin{split}
    &\jsafe = \frac{1}{2} \Big(\qsafehat(s_t, a_t) - 
    (c_t\\ 
    &+ (1 - c_t)\gsafe\underset{a_{t+1} \sim \pi(\cdot | s_{t+1})}{\mathbb{E}}[\qsafehat(s_{t+1}, a_{t+1})])\Big)^2
    \end{split}
\end{align}
W use a target network to create the target values~\cite{learn-to-be-safe,SAC}.

\subsection{Defining a Recovery Set and Policy}
\label{subsec:recovery-region}
\algname executes a composite policy $\pi$ in the environment, which selects between a task-driven policy $\pitask$ and a recovery policy $\pirec$ at each timestep based on whether the agent is in danger of constraint violations in the near future. To quantify this risk, we use $\qsafe$ to construct a recovery set 
that contains state-action tuples from which $\pi$ may not be able to avoid constraint violations.
Then if the agent finds itself in the recovery set, it executes a learned recovery policy instead of $\pitask$ to navigate back to regions of the MDP that are known to be sufficiently safe. Specifically, define two complimentary sets: the safe set $\safeset$ and recovery set $\recset$:
\begin{align*}
    \safeset &= \left\{(s, a)\in\mathcal{S} \times \mathcal{A}: \qsafe(s, a) \leq \esafe \right\}\\
    \recset &= \mathcal{S} \times \mathcal{A}\setminus\safeset
\end{align*}
We consider state-action tuple $(s, a)$ to be safe if in state $s$ after taking action $a$, executing $\pi$ has a discounted probability of constraint violation less than $\esafe$.

If the task policy $\pitask$ proposes an action $a^{\pitask}$ at state $s$ such that $(s, a^{\pitask}) \not \in \safeset$, then a recovery action sampled from $\pirec$ is executed instead of $a^{\pitask}$. Thus, the recovery policy in \algname can be thought of as projecting $\pitask$ into a safe region of the policy space in which constraint violations are unlikely. The recovery policy $\pirec$ is also an RL agent, but is trained to minimize $\qsafehat(s, a)$ to reduce the risk of constraint violations under $\pi$.
Let $a_t^{\pitask}\sim \pitask(\cdot | s_t)$ and $a_t^{\pirec} \sim \pirec(\cdot | s_t)$. Then $\pi$ selects actions as follows:
\begin{align}
\label{eq:composite-policy}
    a_t =
    \begin{cases}
    a_t^{\pitask} & (s_t, a_t^{\pitask}) \in \safeset\\
    a_t^{\pirec} & (s_t, a_t^{\pitask})  \in \recset
    \end{cases}
\end{align}
\algname filters proposed actions that are likely to lead to unsafe states, equivalent to modifying the environment that $\pitask$ operates in with new dynamics:
\begin{align}
    \label{def:new-dynamics}
    P^{\pirec}_{\esafe}(s' | s, a) =
    \begin{cases}
        P(s'| s, a) & (s, a) \in \safeset\\
        P(s'| s, a^{\pirec}) & (s, a) \in \recset
    \end{cases}
\end{align}
We train $\qsafehat$ on samples from $\pi$ since $\pitask$ is not executed directly in the environment, but is rather filtered through $\pi$. 

It is easy to see that the proposed recovery mechanism will shield the agent from regions in which constraint violations are likely if $\qsafehat$ is correct and executing $\pirec$ reduces its value. However, this poses a potential concern: while the agent may be safe, how do we ensure that $\pitask$ can make progress in the \textit{new} MDP defined in equation~\ref{def:new-dynamics}? Suppose that $\pitask$ proposes an unsafe action $a_t^{\pitask}$ under $\qsafehat$. Then, \algname executes a recovery action $a_t^{\pirec}$ and observes transition $(s_t,a_t^{\pirec},s_{t+1}, r_t)$ in the environment. However, if $\pitask$ is updated with this observed transition, it will not learn to associate its proposed action ($a_t^{\pitask}$) in the new MDP with $r_t$ and $s_{t+1}$. As a result, $\pitask$ may continue to propose the same unsafe actions without realizing it is observing the result of an action sampled from $\pirec$. To address this issue, for training $\pitask$, we \textit{relabel all actions with the action proposed by $\pitask$}. Thus, instead of training $\pitask$ with executed transitions $(s_t,a_t,s_{t+1}, r_t)$, $\pitask$ is trained with transitions $(s_t, {\color{red} a_t^{\pitask}}, s_{t+1}, r_t)$. This ties into the interpretation of defining a safe MDP with dynamics $ P^{\pirec}_{\esafe}(s' | s, a)$ for $\pitask$ to act in since all transitions for training $\pitask$ are relabeled as if $\pitask$ was executed directly. 

  \begin{minipage}{0.95\linewidth}
\centering
\begin{algorithm}[H]
\caption{\algname}
\label{alg:main}
\begin{algorithmic}[1]
\Require $\demos$, task horizon $H$, number of episodes $N$
\State Pretrain $\pirec$ and $\qsafehat$ on $\demos$ \Comment{Section~\ref{subsec:offline-pretraining}}
\State $\mathcal{D}_{\mathrm{task}} \leftarrow \emptyset$, $\mathcal{D}_{\mathrm{rec}} \leftarrow \demos$
\State $s_0 \leftarrow \texttt{env.reset()}$
\For{$i \in \{1,\ldots N\}$}
    \For{$t \in \{1,\ldots H\}$}
        \If{$c_t = 1$ or \texttt{is\_terminal}($s_t$)}
            \State $s_t \leftarrow \texttt{env.reset()}$
        \EndIf
        \State $a_t^{\pitask}\sim \pitask(\cdot | s_t)$ \Comment{Query task policy}
        \State\Comment{Check if task policy will be unsafe}
        \If{$(s_t, a_t^{\pitask}) \in \recset$}
            \State $a_t\sim\pirec(\cdot | s_t)$ \Comment{Select recovery policy}
        \Else
            \State $a_t = a_t^{\pitask}$ \Comment{Select task policy}
        \EndIf
        \State Execute $a_t$
        \State Observe $s_{t+1}$, $r_t = R(s_t, a_t)$, $c_t = C(s_t)$
        \State \Comment{Relabel transition}
        \State $\mathcal{D}_{\mathrm{task}}\leftarrow \mathcal{D}_{\mathrm{task}}\cup \{(s_t, a_t^{\pitask}, s_{t+1}, r_t)\}$
        \State $\mathcal{D}_{\mathrm{rec}}\leftarrow \mathcal{D}_{\mathrm{rec}}\cup \{(s_t, a_t, s_{t+1}, c_t)\}$
        \State Train $\pitask$ on $\mathcal{D}_{\mathrm{task}}$, $\pirec$ on $\mathcal{D}_{\mathrm{rec}}$
        \State Train $\qsafehat$ on $\mathcal{D}_{\mathrm{rec}}$ \Comment{Eq.~\ref{eq:safety_critic_loss}}
    \EndFor
\EndFor
\end{algorithmic}
\end{algorithm}
\end{minipage}

\subsection{Offline Pretraining}
\label{subsec:offline-pretraining}
To convey information about constraints before interaction with the environment, we provide the agent with a set of transitions $\demos$ that contain constraint violations for pretraining. While this requires violating constraints in the environment, this data can be collected by human defined policies or under human supervision, and thus provide the robotic agent with examples of constraint violations without the robot having to experience too many uncontrolled examples online. We pretrain $\qsafehat$ by minimizing Equation~\ref{eq:safety_critic_loss} over offline batches sampled from $\demos$. We also pretrain $\pirec$ using $\demos$. Then, $\pitask$, $\pirec$, and $\qsafehat$ are all updated online using experience from the agent's composite policy as discussed in Section~\ref{subsec:recovery-region} and illustrated in Algorithm~\ref{alg:main}. Any RL algorithm can be used to represent $\pitask$ while any off-policy RL algorithm can be used to learn $\pirec$. For some environments in which exploration is challenging, we use a separate set of task demos to initialize $\pitask$ to expedite learning.
\subsection{Practical Implementation}
\label{subsec:implementation}

\textbf{Recovery Policy: } 
Any off-policy RL algorithm can be used to learn $\pirec$. In this paper, we explore both model-free and model-based RL algorithms to learn $\pirec$. For model-free recovery, we perform gradient descent on the safety critic $\qsafehat(s, \pirec(s))$, as in the popular off-policy RL algorithm DDPG~\cite{DDPG}. For model-based recovery, we perform model predictive control (MPC) over a learned dynamics model $f_\theta$ using the safety critic as a cost function. For lower dimensional tasks, we utilize the PETS algorithm from~\citet{handful-of-trials} to plan over a learned stochastic dynamics model, while for tasks with visual observations, we use a VAE based latent dynamics model.
\textbf{Task Policy: } We utilize the popular maximum entropy RL algorithm SAC~\cite{SAC} to learn $\pitask$, but note that any RL algorithm could be used. Details on the implementation of both policies is in the supplement. 

\section{Experiments}
\label{sec:exps}
In the following experiments, we aim to study whether \algname can (1) more effectively trade off task performance and constraint satisfaction than prior algorithms, which jointly optimize both and (2) effectively use offline data for safe RL.

\begin{figure*}[t] %
     \centering
     \includegraphics[width=\textwidth]{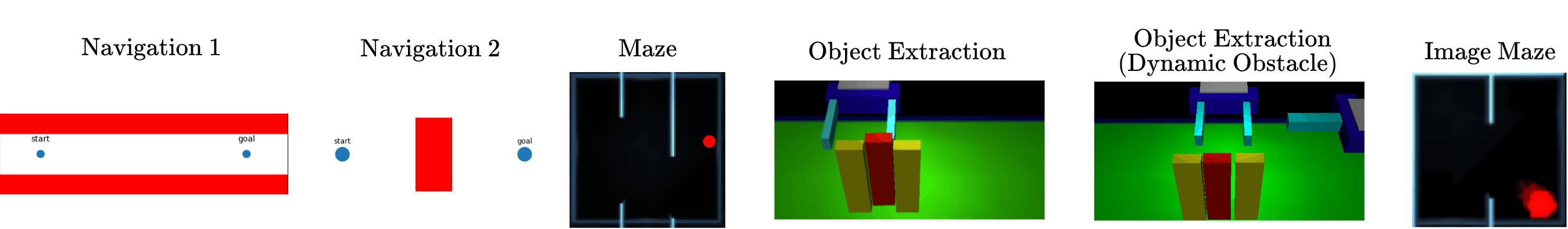}
     \caption{\textbf{Simulation Experiments Domains:} We evaluate \algname on a set of 2D navigation tasks, two contact rich manipulation environments, and a visual navigation task. In Navigation 1 and 2, the goal is to navigate from the start set to the goal set without colliding into the obstacles (red) while in the Maze navigation tasks, the goal is to navigate from the left corridor to the red dot in the right corridor without colliding into walls/borders. In both object extraction environments, the objective is to grasp and lift the red block without toppling any of the blocks or colliding with the distractor arm (Dynamic Obstacle environment).}
        \label{fig:domains}
\end{figure*}

\begin{figure}
     \centering
      \includegraphics[width=0.4\textwidth]{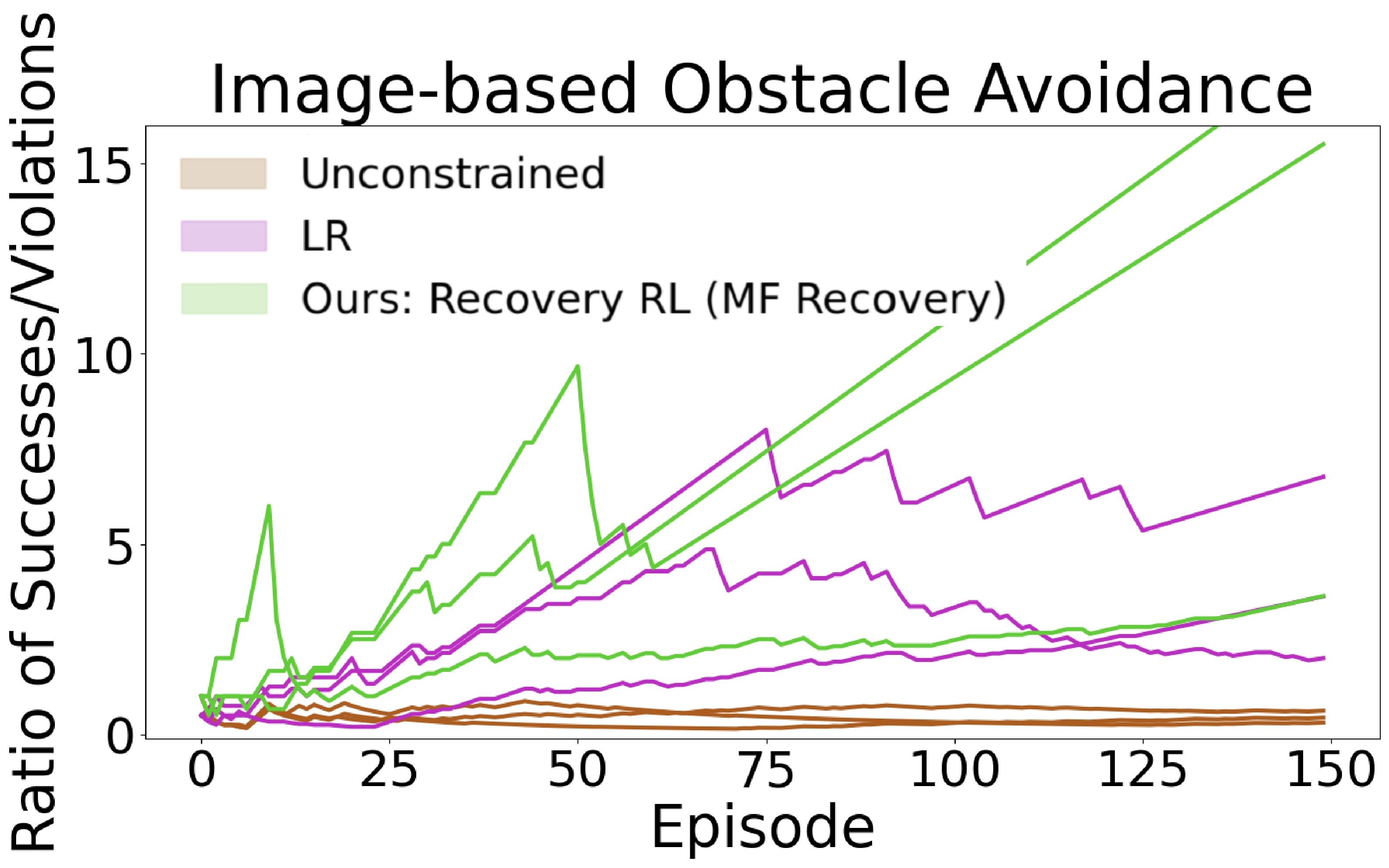}
    \caption{\textbf{Physical Experiment}: We evaluate \algname on an image-based obstacle avoidance task (red obstacles) on the dVRK (Figure~\ref{fig:robot}). We supply all algorithms with an overhead RGB image as input and run each algorithm 3 times. We find that \algname significantly outperforms Unconstrained and LR.}
    \label{fig:reacher}
\end{figure}

\textbf{Domains:}
We evaluate \algname on a set of 6 simulation domains (Figure~\ref{fig:domains}) and an image-based obstacle avoidance task on a physical robot (Figure~\ref{fig:reacher}). 
All experiments involve policy learning under state space constraints, in which a constraint violation terminates the current episode. This makes learning especially challenging, since constraint violations directly preclude further exploration. This setting is reflective of a variety of real world environments, in which constraint violations can require halting the robot due to damage to itself or its surrounding environment.

We first consider three 2D navigation domains: Navigation 1, Navigation 2, and Maze. Here, the agent only observes its position in 2D space and experiences constraint violations if it hits obstacles, walls, or workspace boundaries. We then consider three higher dimensional tasks to evaluate whether \algname can be applied to contact rich manipulation tasks (Object Extraction, Object Extraction (Dynamic Obstacle)) and vision-based continuous control (Image Maze). In the object extraction environments, the goals is to extract the red block without toppling any blocks, and in the case of Object Extraction (Dynamic Obstacle), also avoiding contact with a dynamic obstacle which moves in and out of the workspace. Image Maze is a shorter horizon version of Maze, but the agent is only provided with image observations rather than its $(x,y)$ position in the environment.

We then evaluate \algname on an image-based obstacle avoidance task on the da Vinci Research Kit (dVRK)~\cite{kazanzides-chen-etal-icra-2014} where the robot must guide its end effector within 2 mm of a target position from two possible starting locations without touching red 3D printed obstacles in the workspace. See Figure~\ref{fig:robot} for an illustration of the experimental setup. The dVRK is cable-driven and has relatively imprecise controls, motivating closed-loop control strategies to compensate for these errors~\cite{hwang2020efficiently}. Furthermore, the dVRK system has been used in the past to evaluate safe RL algorithms~\cite{SAVED} due to its high cost and the delicate structure of its arms, which make safe learning critical. Further environment, task, and data collection details can be found in the supplement for all simulation and physical experiments.

\textbf{Offline Data Collection: }
To effectively initialize $\qsafehat$, $\demos$ should ideally contain a diverse set of trajectories which violate constraints in different ways. Since $\demos$ need not be task specific, data from other tasks in the environment could be used, or simple human defined policies can be used to illustrate constraint violating behaviors. We take the latter approach: for all navigation environments (Navigation 1, Navigation 2, Maze, Image Maze, and the physical experiment), offline data is collected by initializing the agent in various regions of the environment and directing the agent towards the closest obstacle. For the object extraction environments (Object Extraction, Object Extraction (Dynamic Obstacle)), demonstrations are collected by guiding the end effector towards the target red block and adding Gaussian noise to controls when it is sufficiently close to the target object to make toppling likely. \algname and all comparisons which have a safety critic are given the same offline dataset $\demos$. See the supplementary material for details on the data collection procedure, and the number of total transitions and constraint violating states for all offline datasets.

\textbf{Evaluation Metric:}
Since \algname and prior methods trade off between safety and task progress, we report the ratio of the cumulative number of task successes and the cumulative number of constraint violations at each episode to illustrate this (higher is better). We tune all algorithms to maximize this ratio, and task success is determined by defining a goal set in the state space for each environment. To avoid issues with division by zero, we add 1 to the cumulative task successes and constraint violations when computing this ratio. This metric provides a single scalar value to quantify how efficiently different algorithms balance task completion and constraint satisfaction. We do not report reward per episode, as episodes terminate on task completion or constraint violation. Each run for simulation experiments is replicated across $10$ random seeds and we report the mean and standard error. For physical experiments we run each algorithm across $3$ random seeds and visualize all 3 runs. In the supplementary material, we also report additional metrics for each experiment: cumulative task successes, cumulative constraint violations, and reward learning curves. We find that \algname violates constraints less often than comparisons while maintaining a similar task success rate and more efficiently optimizing the task reward.

\begin{figure*}
     \centering
     \includegraphics[width=\textwidth]{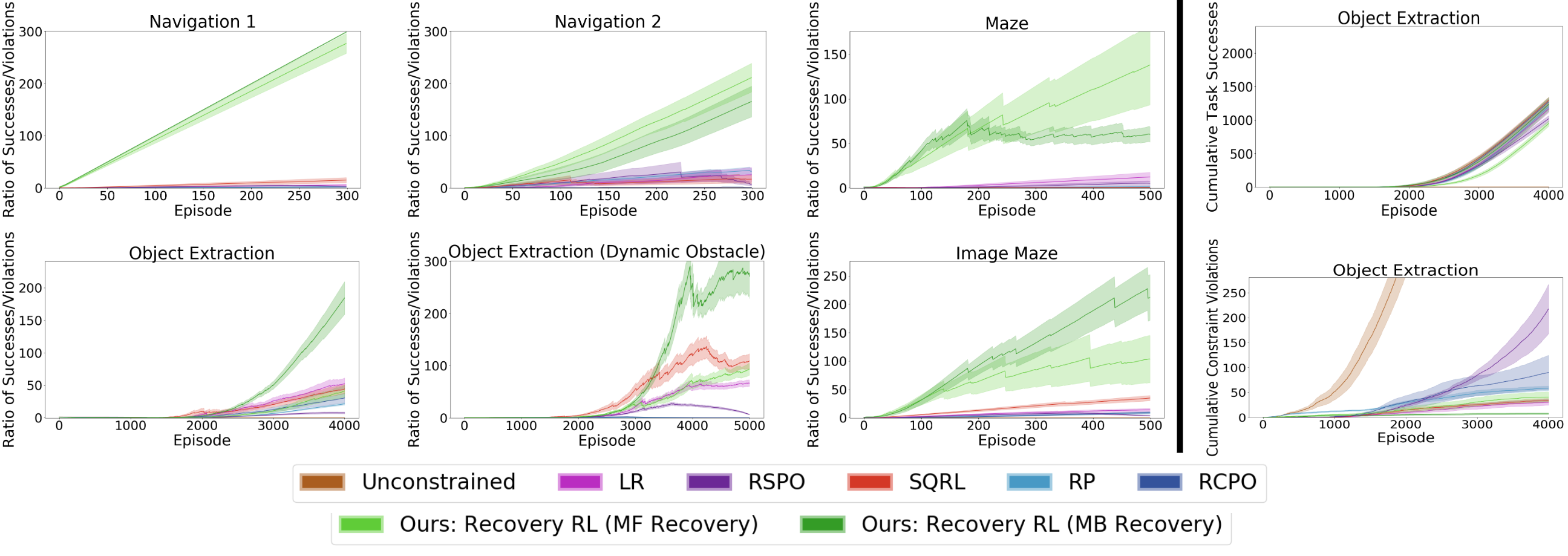}
     \caption{\textbf{Simulation Experiments: }\underline{Left: ratio of successes to constraint violations over the course of online training.} In all navigation tasks, we find that \algname significantly outperforms prior methods with both model-free and model-based recovery policies, while for the object extraction environments, \algname with a model-based recovery policy significantly outperforms prior algorithms while \algname with a model-free recovery policy does not perform as well. We hypothesize that this is due to the model-based recovery mechanism being better able to compensate for imperfections in $\qsafehat$. Results are averaged over 10 runs for each algorithm; the sawtooth pattern occurs due to constraint violations, which result in a sudden drop in the ratio. \underline{Right: cumulative successes and constraint violations.} Additionally, we show the cumulative task successes and cumulative constraint violations for the Object Extraction task for all algorithms, and find that \algname with model-based recovery succeeds more often than all comparisons while also violating constraints the least. Similar plots for all other experimental domains can be found in the supplementary material.}
        \label{fig:sim-exps}
\end{figure*}
\textbf{Comparisons:}
\label{subsec:comparisons}
We compare \algname to the following algorithms that ignore constraints (Unconstrained) or enforce constraints via the optimization objective (LR, SQRL, RSPO) or via reward shaping (RP, RCPO). 
\begin{itemize}[topsep=0pt,
leftmargin=*]
    \item \textbf{Unconstrained}: optimizes task reward, ignoring constraints.
    \item \textbf{Lagrangian Relaxation (LR)}: minimizes $L_{\mathrm{policy}}(s, a, r, s';\pi) + \lambda (\mathbb{E}_{a \sim \pi(\cdot | s)}\left[\qsafehat(s, a)\right] - \esafe)$, where $L_{\mathrm{policy}}$ is the policy optimization loss and the second term approximately enforces $\qsafehat(s,a) \leq \esafe$. Policy parameters and $\lambda$ are updated via dual gradient descent.
    \item \textbf{Safety Q-Functions for RL (SQRL)~\cite{learn-to-be-safe}}: combines the LR method with a filtering mechanism to reject policy actions for which $\qsafehat(s, a) > \esafe$.
    \item \textbf{Risk Sensitive Policy Optimization (RSPO)}~\cite{risk-sensitive-RL}: minimizes $L_{\mathrm{policy}}(s, a, r, s';\pi) + \lambda_t (\mathbb{E}_{a \sim \pi(\cdot | s)}\left[\qsafehat(s, a)\right] - \esafe)$, where $\lambda_t$ is a sequence which decreases to $0$.
    \item \textbf{Reward Penalty (RP)}: observes a reward function that penalizes constraint violations: $R'(s,a) = R(s,a) - \lambda C(s)$.
    \item \textbf{Critic Penalty Reward Constrained Policy Optimization (RCPO)~\cite{RCPO}}: optimizes the Lagrangian relaxation via dual gradient descent and the policy gradient trick. The policy gradient update maximizes $\mathbb{E}_\pi\left[\sum_{t=0}^\infty \gamma^t (R(s_t,a_t) - \lambda \qsafehat(s_t, a_t)) \right]$ and the multiplier update is the same as in LR.
\end{itemize}

\begin{figure*}
     \centering
     \includegraphics[width=\textwidth]{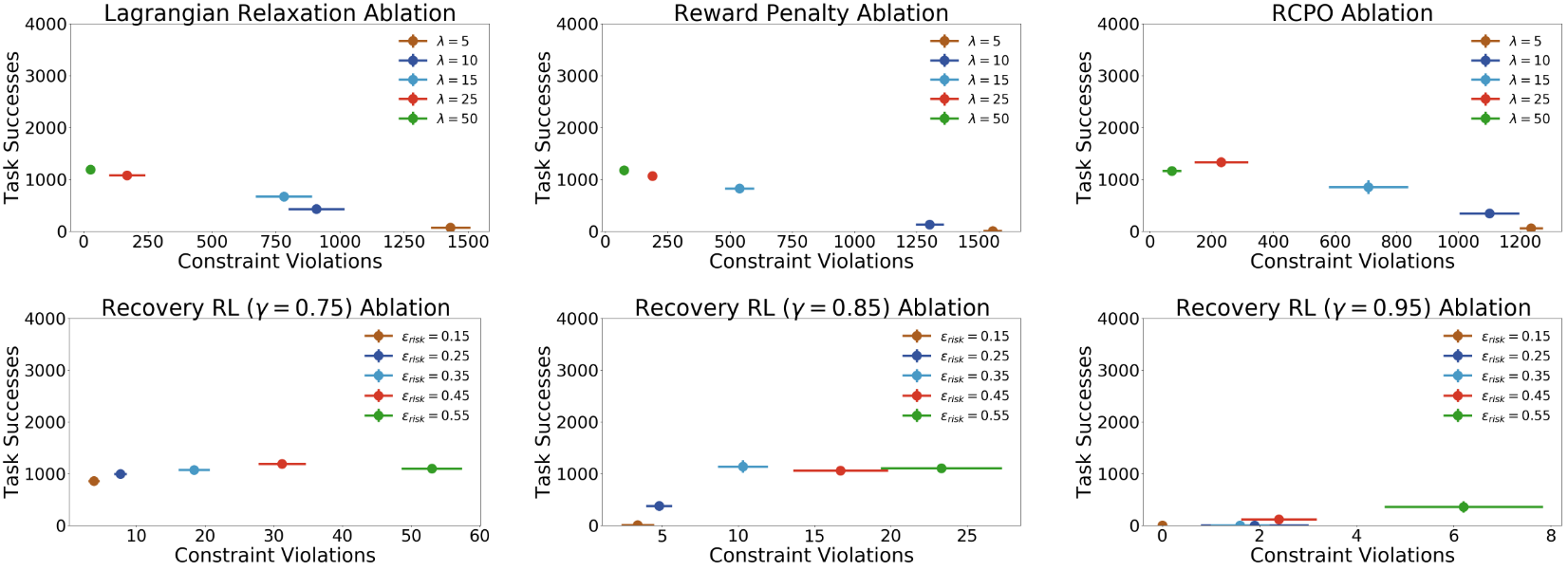}
     \caption{\textbf{Sensitivity Experiments: } We report the final number of task successes and constraint violations averaged over 10 runs at the end of training for \algname and comparison algorithms for a variety of different hyperparameter settings on the Object Extraction task. We find that the comparison algorithms are relatively sensitive to the value of the penalty parameter $\lambda$ while given a fixed $\gsafe$, \algname achieves relatively few constraint violations while maintaining task performance over a range of $\esafe$ values.}
     \label{fig:sensitivity_exps}
\end{figure*}

All of these algorithms are implemented with the same base algorithm for learning the task policy (Soft Actor-Critic~\cite{SAC}) and all but Unconstrained and RP are modified to use the same safety critic $\qsafehat$ which is pretrained on $\demos$ for all methods. Thus, the key difference between \algname and prior methods is how $\qsafehat$ is utilized: the comparisons use a joint objective which uses $\qsafehat$ to train a single policy that optimizes for both task performance and constraint satisfaction, while \algname separates these objectives across two sub-policies. We tune all prior algorithms and report the best hyperparameter settings found on each task for the ratio-based evaluation metric. Details on \algname and all comparison algorithms are in the supplement.

\textbf{Results: }We first study the performance of \algname and prior methods in all simulation domains in Figure~\ref{fig:sim-exps}. Results suggest that \algname with both model-free and model-based recovery mechanisms significantly outperform prior algorithms across all 3 2D pointmass navigation environments (Navigation 1, Navigation 2, Maze) and the visual navigation environment (Image Maze). In the Object Extraction environments, we find that \algname with model-based recovery significantly outperforms prior algorithms, while \algname with a model-free recovery mechanism does not perform nearly as well. We hypothesize that the model-based recovery mechanism is better able to compensate for approximation errors in $\qsafehat$, resulting in a more robust recovery policy. We find that the prior methods often get very low ratios since they tend to achieve a similar number of task completions as \algname, but with many more constraint violations. In contrast, \algname is generally able to effectively trade off between task performance and safety. This is illustrated on the right pane of Figure~\ref{fig:sim-exps}, which suggests that \algname with model-based recovery not only succeeds more often than comparison algorithms, but also exhibits fewer constraint violations. We study this further in the supplement. Finally, we evaluate \algname and prior algorithms on the image-based obstacle avoidance task illustrated in Figure~\ref{fig:robot} and find that \algname substantially outperforms prior methods, suggesting that \algname can be used for contact-rich visuomotor control tasks in the real world (Figure~\ref{fig:reacher}). We study when \algname violates constraints in the supplement, and find that in most tasks, the recovery policy is already activated when constraint violations occur. This is encouraging, because if a recovery policy is challenging to learn, \algname could still be used to query a human supervisor for interventions.
\begin{figure}
     \centering
     \includegraphics[width=0.48\textwidth]{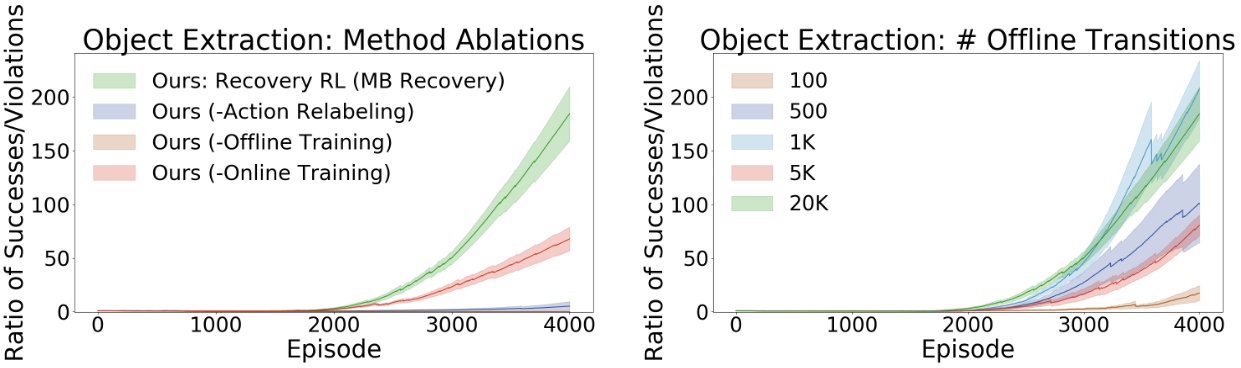}
     \caption{\textbf{Ablations:} We first study the affect of different algorithmic components of \algname (left). Results suggest that offline pretraining of $\pirec$ and $\qsafehat$ is critical for good performance, while removing online updates leads to a much smaller reduction in performance. Furthermore, we find that the action relabeling method for training $\pitask$ (Section~\ref{subsec:recovery-region}) is critical for good performance. We then study the sensitivity of \algname to the number of offline transitions used to pretrain $\pirec$ and $\qsafehat$ (right) and find that \algname performs well even with just $1000$ transitions in $\demos$ for the Object Extraction task, with performance degrading when the number of transitions is reduced beyond this point.}
        \label{fig:ablations}
\end{figure}

\textbf{Ablations: }We ablate different components of \algname and study the sensitivity of \algname to the number of transitions in $\demos$ for the Object Extraction domain in Figure~\ref{fig:ablations}. Results suggest that \algname performs much more poorly when $\pirec$ and $\qsafehat$ are not pretrained with data from $\demos$, indicating the value of learning to reason about safety before environment interaction. However, when $\pirec$ and $\qsafehat$ are not updated online, performance degrades much less significantly. A key component of \algname is relabeling actions when training the task policy so that $\pitask$ can learn to associate its proposed actions with their outcome (Section~\ref{subsec:recovery-region}). We find that without this relabeling, \algname achieves very poor performance as it rarely achieves task successes. Additionally, we find that although the reported simulation experiments supply \algname and all prior methods with $20,000$ transitions in $\demos$ for the Object Extraction task, \algname is able to achieve good performance with just $1000$ transitions in $\demos$, with performance significantly degrading only when the size of $\demos$ is reduced to less than this amount.

\textbf{Sensitivity Experiments: }We tune hyperparameters for \algname and all baselines to ensure a fair comparison. We first tune $\gsafe$ and $\esafe$ for \algname, and then use the same $\gsafe$ and $\esafe$ for prior methods to ensure that all algorithms use the same safety critic training procedure. These two hyperparameters are the only ones tuned for \algname and SQRL. For RP, RCPO, and LR, we tune the penalty term $\lambda$ with $\gsafe$ and $\esafe$ fixed as mentioned above. For RSPO, we utilize a schedule which decays $\lambda$ from 2 times the best value found for $\lambda$ when tuning the LR comparison to 0 with an evenly spaced linear schedule over all training episodes. In Figure~\ref{fig:sensitivity_exps}, we study the sensitivity of \algname with model-based recovery and the RP, RCPO, and LR comparisons to different hyperparameter choices on the Object Extraction task. \algname appears less sensitive to hyperparameters than the comparisons for the $\gsafe$ values we consider.
\vspace{-0.1in}

\section{Conclusion}
\label{sec:conclusion}
We present \algname, a new algorithm for safe RL which is able to more effectively balance task performance and constraint satisfaction than 5 state-of-the-art prior algorithms for safe RL across 6 simulation domains and an image-based obstacle avoidance task on a physical robot. In future work we hope to explore further evaluation on physical robots, establish formal guarantees, and use ideas from offline RL to more effectively pretrain the recovery policy. We will explore settings in which constraint violations may not be catastrophic and applications for large-scale robot learning.
\vspace{-0.1in}

\section{Acknowledgments}
\footnotesize
This research was performed at the AUTOLAB at UC Berkeley in affiliation with the Berkeley AI Research (BAIR) Lab, the Real-Time Intelligent Secure Execution (RISE) Lab, Google Brain Robotics, and the Stanford AI Research Lab. Authors were also supported by the SAIL-Toyota Research initiative, the Scalable Collaborative Human-Robot Learning (SCHooL) Project, the NSF National Robotics Initiative Award 1734633, and in part by donations from Google, Siemens, Amazon Robotics, Toyota Research Institute, and by equipment grants from NVidia. This article solely reflects the opinions and conclusions of its authors and not the views of the sponsors or their associated entities. A.B. and S.N. are supported by NSF GRFPs. We thank our colleagues who provided helpful feedback, code, and suggestions, in particular Jie Tan, Jeffrey Ichnowski, and Wisdom Agboh. 
\renewcommand*{\bibfont}{\footnotesize}
\renewcommand{\baselinestretch}{0.9} %
\printbibliography %
\clearpage
\twocolumn[
\begin{LARGE}
\begin{center}
\textbf{Recovery RL: Safe Reinforcement Learning with Learned Recovery Zones Supplementary Material}
\end{center}
\end{LARGE}]
The supplementary material is structured as follows: In Section~\ref{sec:recovery-rl-discussion} we discuss brief theoretical motivation for \algname and possible variants and in Section~\ref{sec:alg-details} we discuss algorithmic details for \algname and comparison algorithms. In Section~\ref{sec:metrics}, we report additional metrics for all domains and comparisons and in Section~\ref{sec:safety-critic-vis} we visualize the safety critic for all navigation experiments. We provide additional details about algorithm implementation in Section~\ref{sec:imp-details}, and on domain-specific algorithm hyperparameters in Section~\ref{subsec:alg-params}. Finally, we report simulation and physical environment details in Section~\ref{sec:exp-details}.

\section{\algname Theoretical Motivation and Variants}
\label{sec:recovery-rl-discussion}
In this section, we will briefly and informally discuss additional properties of \algname and then discuss some variants of \algname.

\subsection{Theoretical Motivation}
\label{subsec:theory}
Recall that the task policy is operating in an environment with modified dynamics:
\begin{align}
    P^{\pirec}_{\esafe}(s' | s, a) =
    \begin{cases}
        P(s'| s, a) & (s, a) \in \safeset\\
        P(s'| s, a^{\pirec}) & (s, a) \in \recset
    \end{cases}
\end{align}

However, $P^{\pirec}_{\esafe}$ changes over time (even within the same episode) and analysis of policy learning in non-stationary MDPs is currently challenging and ongoing work. Assuming that $P^{\pirec}_{\esafe}$ is stationary following the pretraining phase, it is immediate that $\pitask$ is operating in a stationary MDP $\mathcal{M}'=(\mathcal{S}, \mathcal{A}, P^{\pirec}_{\esafe}, R(\cdot, \cdot), \gamma, \mu)$, and therefore all properties of $\pitask$ in stationary MDPs apply in $\mathcal{M}'$. Observe that iterative improvement for $\pitask$ in $\mathcal{M}'$ implies iterative improvement for $\pi$ in $\mathcal{M}$, since both MDPs share the same reward function, and an action taken by $\pitask$ in $\mathcal{M}'$ is equivalent to $\pitask$ trying the action in $\mathcal{M}$ before being potentially caught by $\pirec$.

\subsection{Safety Value Function}\label{subsec:safety-value}
One variant of \algname can use a safety critic that is a state-value function $\vsafe(s)$ instead of a state-action-value function. While this implementation is simpler, the $\qsafe$ version used in the paper can switch to a safe action instead of an unsafe one instead of waiting to reach an unsafe state to start recovery behavior.

\subsection{Reachability-based Variant}
\label{subsec:reachability}
Another variant can use the learned dynamics model in the model-based recovery policy to perform a one (or $k$) step lookahead to see if future states-action tuples are in $\safeset$. While $\qsafe$ in principle carries information about future safety, this is an alternative method to check future states.
\section{ Algorithm Details}
\label{sec:alg-details}
\subsection{Recovery RL}
\label{subsec:recovery-rl}
\textbf{Recovery Policy: }In principle, any off-policy reinforcement learning algorithm can be used to learn the recovery policy $\pirec$. In this paper, we explore both model-free and model-based reinforcement learning algorithms to learn $\pirec$. For model-free recovery, we perform gradient descent on the safety critic $\qsafehat(s, \pirec(s))$, as in the popular off-policy reinforcement learning algorithm DDPG~\cite{DDPG}. We choose the DDPG-style objective function over alternatives since we do not wish the recovery policy to explore widely. For model-based recovery, we perform model predictive control (MPC) over a learned dynamics model $f_\theta$ by minimizing the following objective:
\begin{align}
    \label{eq:MB-recov-objective}
    L_\theta(s_t, a_t) = \mathbb{E}\left[\sum_{i=0}^H \qsafehat(\hat{s}_{t+i}, a_{t+i})\right]
\end{align}
where $\hat{s}_{t+i+1}\sim f_\theta (\hat{s}_{t+i}, a_{t+i})$, $\hat{s}_t=s_t$, and $\hat{a}=a_t$.  
For lower dimensional tasks, we utilize the PETS algorithm from~\citet{handful-of-trials} to plan over a learned stochastic dynamics model while for tasks with visual observations, we utilize a VAE based latent dynamics model.
In the offline pretraining phase, when model-free recovery is used, batches are sampled sequentially from $\demos$ and each batch is used to (1) train $\qsafehat$ and (2) optimize the DDPG policy to minimize the current $\qsafehat$. When model-based recovery is used, the data in $\demos$ is first used to learn dynamics model $f_\theta$ using either PETS (low dimensional tasks) or latent space dynamics (image-based tasks). Then, $\qsafehat$ is separately optimized to over batches sampled from $\demos$. During the online RL phase, all methods are updated online using on-policy data from composite policy $\pi$.

\textbf{Task Policy: }In experiments, we utilize the popular maximum entropy RL algorithm SAC~\cite{SAC} to learn $\pitask$, but note that any RL algorithm could be used to train $\pitask$. In general $\pitask$ is only updated in the online RL phase. However, in certain domains where exploration is challenging, we pre-train SAC on a small set of task-specific demonstrations to expedite learning. To do this, like for training the model-free recovery policy, we sample batches sequentially from $\demos$ and each batch is used to (1) train $\qsafehat$ and (2) optimize the SAC policy to minimize the current $\qsafehat$. To ensure that $\pitask$ learns which actions result in recovery behavior, we train $\pitask$ on transitions $(s_t, a_t^{\pitask}, s_{t+1})$ even if $\pirec$ was executed.

\subsection{Unconstrained}
\label{subsec:unconstrained}
We use an implementation of the popular model-free reinforcement learning algorithm Soft Actor Critic~\cite{SAC_github, SAC}, which maximizes a combination of task reward and policy entropy with a stochastic actor function.

\subsection{Lagrangian Relaxation (LR)}
\label{subsec:lagrangian}
In this section we will briefly motivate and derive the Lagrangian relaxation comparison. As before, we desire to solve the following constrained optimization problem:
\begin{align*}
    \min_{\pi} L_{\mathrm{policy}}(s; \pi) \textrm{ s.t. } \mathbb{E}_{a\sim\pi(\cdot|s)}\left[\qsafe(s, a)\right] \leq \esafe
\end{align*}
where $L_{\mathrm{policy}}$ is a policy loss function we would like to minimize (e.g. from SAC). As in prior work in solving constrained optimization problems, we can solve the following unconstrained problem instead:
\begin{align*}
    \max_{\lambda \geq 0}\min_{\pi} L_{\mathrm{policy}}(s; \pi) + \lambda (\mathbb{E}_{a\sim\pi(\cdot|s)}\left[\qsafe(s, a)\right] - \esafe )
\end{align*}
We aim to find a saddle point of the Lagrangian function via dual gradient descent. In practice, we use samples to approximate the expectation in the objective by sampling an action from $\pi(\cdot|s)$ each time the objective function is evaluated.

\subsection{Risk Sensitive Policy Optimization (RSPO)}
\label{subsec:RSPO}
We implement Risk Sensitive Policy Optimization by implementing the Lagrangian Relaxation method as discussed in Section~\ref{subsec:lagrangian} with a sequence of multipliers which decrease over time. This encourages initial constraint satisfaction followed by gradual increase in prioritization of the task objective and is inspired by the Risk Sensitive Q-learning algorithm from~\cite{risk-sensitive-RL}.

\subsection{Safety Q-Functions for Reinforcement Learning (SQRL)}
\label{subsec: SQRL}
This comparison is identical to LR, except it additionally adds a Q-filter, that performs rejection sampling on the policy's distribution $\pi(\cdot|s_t)$ until it finds an action $a_t$ such that $\qsafe(s_t, a_t) \leq \esafe$.

\subsection{Reward Penalty (RP)}
\label{subsec:RP}
The reward penalty comparison simply involves subtracting a constant penalty $\lambda$ from the task reward function when a constraint is violated. This is the only comparison algorithm other than Unconstrained which does not use the learned $\qsafe$ or the constraint demos, but is included due to its surprising efficacy and simplicity.

\subsection{Off Policy Reward Constrained Policy Optimization (RCPO)}
\label{subsec:RCPO}
In on-policy RCPO~\cite{RCPO}, the policy is optimized via policy gradient estimators by maximizing $\mathbb{E}_{\pi}\left[\sum_{t=0}^\infty\left( \gamma^t R(s, a) - \lambda \gsafe^t D(s, a)\right)\right]$. In this work, we use $D(s, a) =  \qsafe(s,a)$ and update the Lagrange multiplier $\lambda$ as in LR. We could also use $D(s,a) = C(s)$, which would be almost identical to the RP comparison. Instead of optimizing this with on-policy RL, we use SAC to optimize it in an off-policy fashion to be consistent with the other comparisons.

\section{Additional Experimental Metrics}
\label{sec:metrics}
\begin{figure*}
     \centering
     \includegraphics[width=\textwidth]{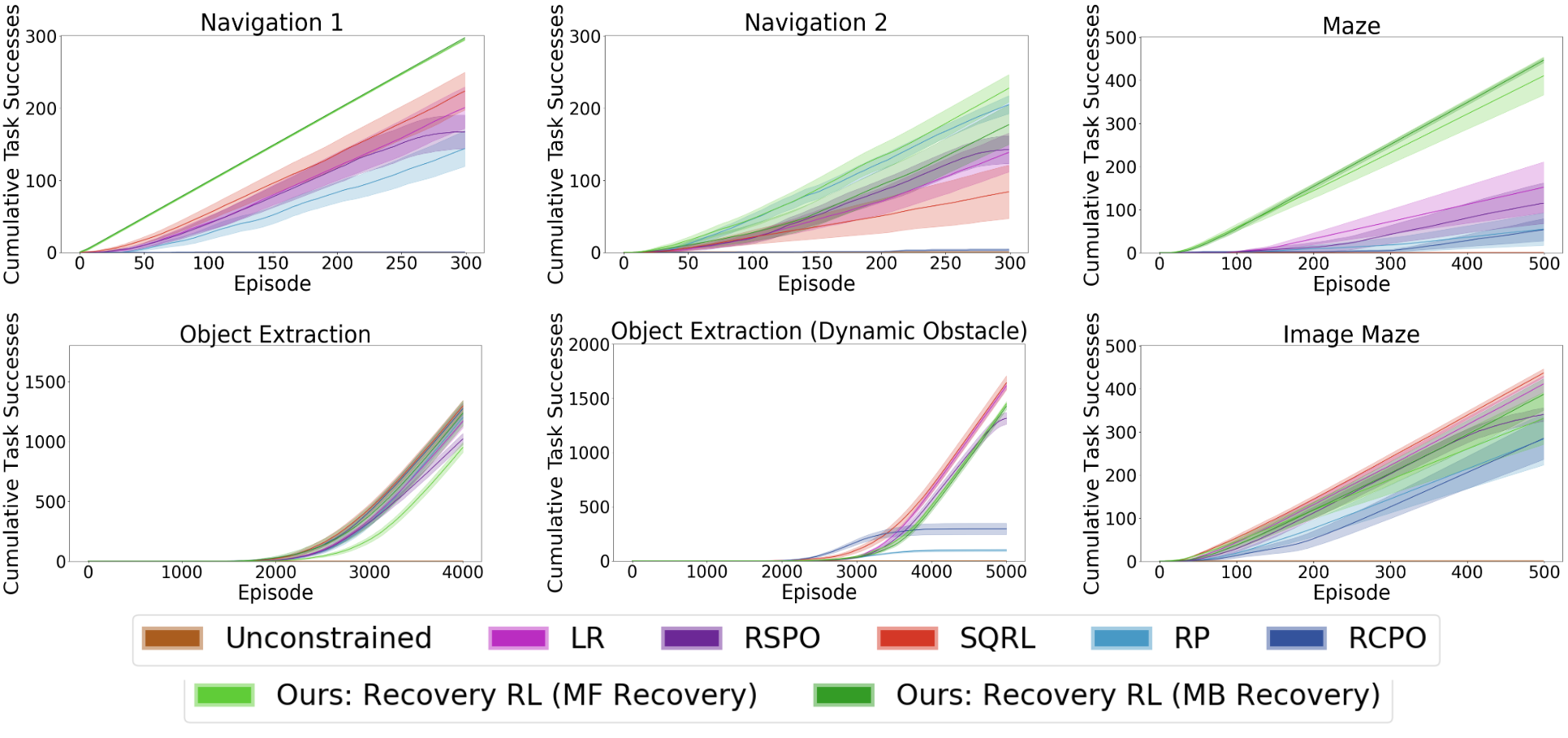}
     \caption{\textbf{Simulation Experiments Cumulative Successes: }We plot the cumulative task successes for each algorithm in each simulation domain, with results averaged over 10 runs for all algorithms. We observe that \algname (green), is generally among the most successful algorithms. In the cases that it has lower successes, we observe that it is safer (Figure~\ref{fig:sim-exps-violations}). We find that \algname has a higher or comparable task success rate to the next best algorithm on all environments except for the Object Extraction (Dynamic Obstacle) environment.}
        \label{fig:sim-exps-successes}
\end{figure*}

\begin{figure*}
     \centering
     \includegraphics[width=\textwidth]{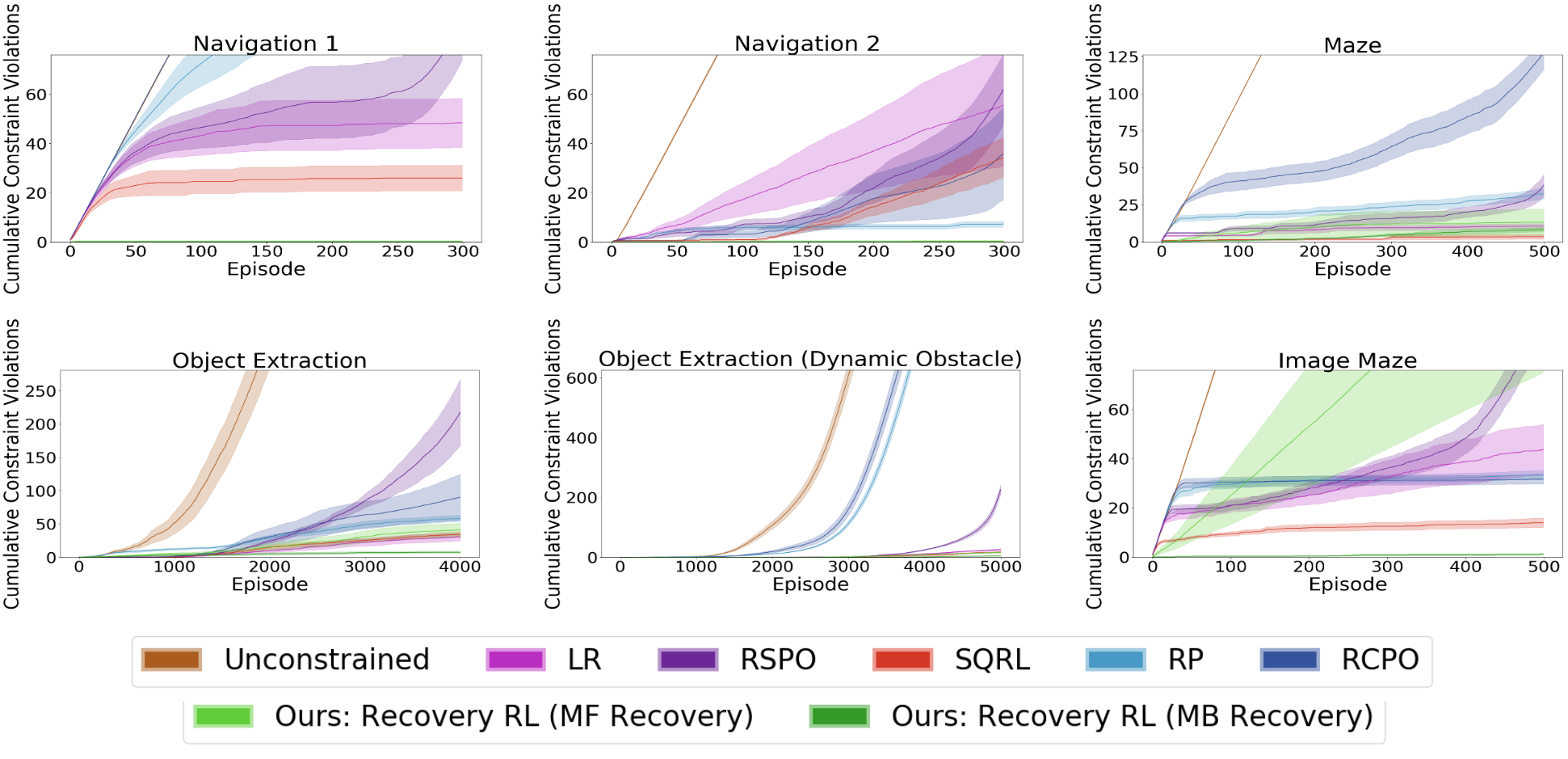}
     \caption{\textbf{Simulation Experiments Cumulative Violations: }We plot the cumulative constraint violations for each algorithm in each simulation domain, with results averaged over 10 runs for all algorithms. We observe that \algname (green), is among the safest algorithms across all domains. In the cases where it is less safe than a comparison, it has a higher task success rate (Figure~\ref{fig:sim-exps-successes}).}
        \label{fig:sim-exps-violations}
\end{figure*}

\begin{figure*}
     \centering
     \includegraphics[width=\textwidth]{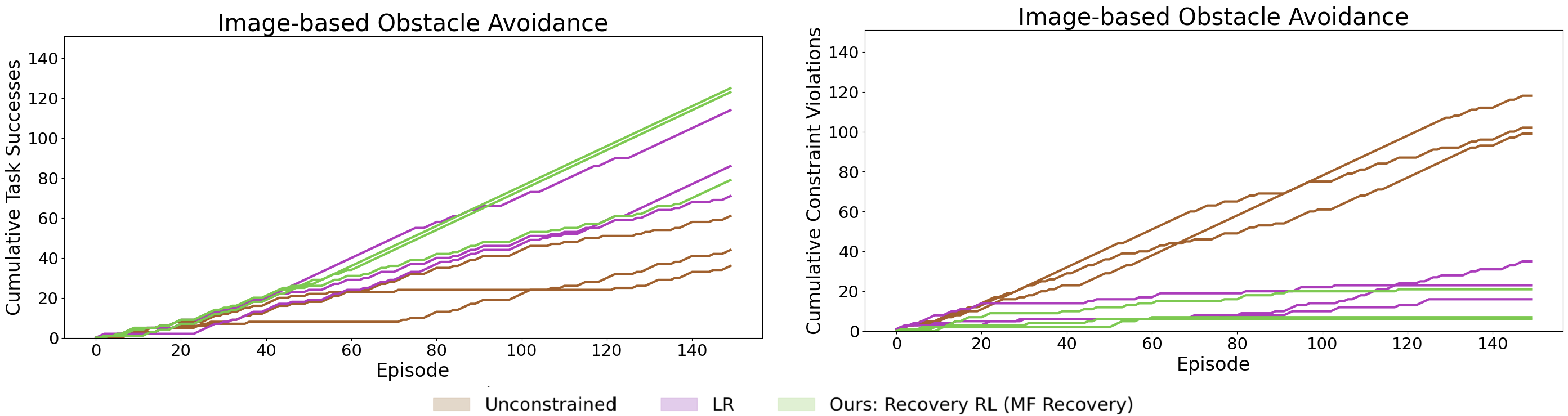}
     \caption{\textbf{Physical Experiment Successes and Violations: }We plot the cumulative constraint violations and task successes for the image-based obstacle avoidance task on the dVRK for all 3 runs of each algorithm. We observe that \algname is generally both more successful and safer than LR and unconstrained.}
        \label{fig:phys-exps-stats}
\end{figure*}

In Figure~\ref{fig:sim-exps-successes} and Figure~\ref{fig:sim-exps-violations}, we report cumulative task successes and constraint violations for all methods for all simulation experiments. We report these statistics for the image-based obstacle avoidance physical experiment in Figure~\ref{fig:phys-exps-stats}. We observe that \algname is generally very successful across most domains with relatively few violations. Some more successful comparisons tend to have many more constraint violations.

\begin{figure*}
     \centering
     \includegraphics[width=\textwidth]{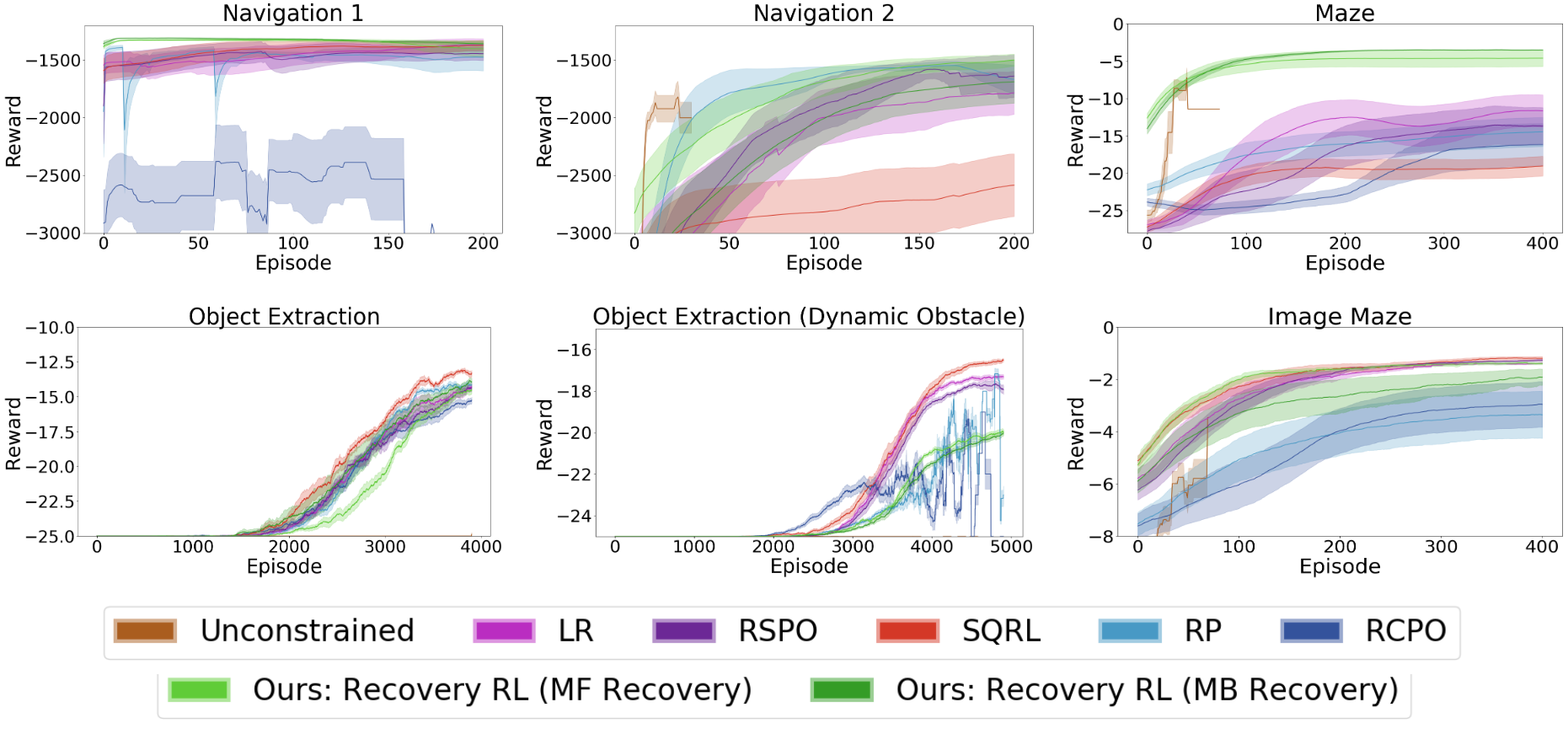}
     \caption{\textbf{Simulation Experiments Reward Learning Curve: }We show the total reward attained in each episode smoothed over a 100 episode length window for each simulation domain, with results averaged over 10 runs for all algorithms. We do not show results when constraints are violated to prevent negative bias for algorithms which may violate constraints very frequently, which accounts for gaps in the plot, especially for the unconstrained algorithm which tends to violate constraints very frequently. Thus, the plots illustrate the attained reward for all algorithms conditioned on not violating constraints, which provides a good measure on the quality of solutions found. We find that on all but the dynamic obstacle domain, \algname is able to converge more quickly to higher quality solutions with respect to the task reward function compared to comparisons. This indicates that \algname is able to learn more efficiently, in addition to more safely, compared to comparison algorithms.}
        \label{fig:sim-exps-rewards}
\end{figure*}

\begin{figure}
     \centering
     \includegraphics[width=0.47\textwidth]{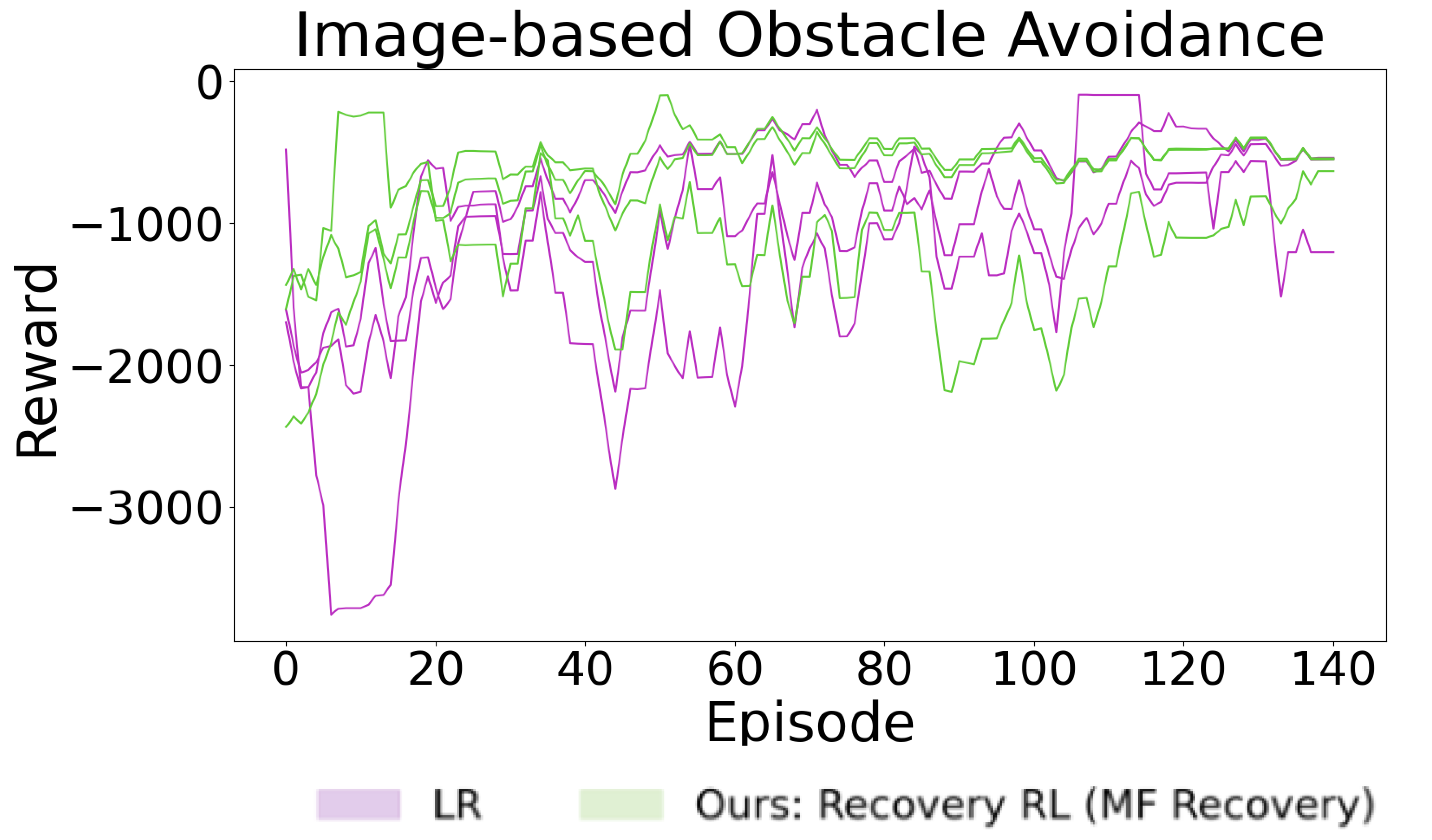}
     \caption{\textbf{Physical Experiment Reward Learning Curve: }We show the total reward attained in each episode smoothed over a 10 episode length window with results from 3 runs for all algorithms. We do not show results when constraints are violated to prevent negative bias for algorithms which may violate constraints very frequently, which accounts for gaps in the plot or explains why some plots have a single line (only one out of 3 runs is constraint satisfying) . Note that the Unconstrained algorithm does not appear in the plot as it never makes progress on the harder initial configuration of the task. Thus, the plots illustrate the attained reward for all algorithms conditioned on not violating constraints, which provides a good measure on the quality of solutions found. We find that \algname is able to converge more quickly to higher quality solutions with respect to the task reward function compared to comparisons. This indicates that \algname is able to learn more efficiently, in addition to more safely, compared to comparison algorithms.}
        \label{fig:phys-exps-rewards}
\end{figure}

Additionally, in Figures~\ref{fig:sim-exps-rewards} and~\ref{fig:phys-exps-rewards}, we plot the cumulative reward attained by the agent for \algname and all comparison algorithms to evaluate whether \algname learns more efficiently than comparisons while also learning safely. For all plots, we show total reward attained in each episode smoothed over a 100 episode length window. Additionally, we do not show results when constraints are violated to prevent negative bias for algorithms which may violate constraints very frequently, which accounts for gaps in the plot, especially for the unconstrained algorithm which tends to violate constraints very frequently. Thus, the plots illustrate the attained reward for all algorithms conditioned on not violating constraints, which provides a good measure on the quality of solutions found. We find that in addition to the safe learning shown by \algname as evidenced by the results in Figure~\ref{fig:sim-exps-violations}, the results in Figure~\ref{fig:sim-exps-rewards} and Figure~\ref{fig:phys-exps-rewards}  indicate that when \algname satisfies constraints, it generally converges to higher quality solutions more quickly compared to the comparison algorithms. These results provide further evidence that the way \algname separates the often conflicting objectives of task directed optimization and constraint satisfaction allows it to not only be safer during learning, but also learn more efficiently.

In Table~\ref{table:violations}, we report empirical results for when constraint violations occur in Table~\ref{table:violations}. Results suggest that in most tasks, the recovery policy is already activated when violations do occur. Thus, in these failure cases, \algname is able to successfully predict future violations, but is not able to prevent them. This is encouraging, and suggests that for environments in which a recovery policy is very challenging to learn, \algname could still be used to query a human supervisor for interventions.

\begin{table*}[t]
\caption {\textbf{Constraint Violations Breakdown: } We report the proportion of constraint violations for each environment that occur when the recovery policy is activated for \algname (format is mean $\pm$ standard error). If most constraint violations occur when the recovery policy is active, this indicates that the safety critic is sufficiently accurate to detect that the recovery policy must be activated, but may not provide sufficient information to avoid constraint violations. We note that if the safety critic detects the need for recovery behavior too late, then these errors are attributed to the recovery policy. For the both Maze environments and the Object Extraction environment, most constraint violations occur when the recovery policy is activated. In Navigation 1, none occur when the recovery policy is activated, but in this environment constraints are almost never violated. In the Image-Based obstacle avoidance tasks, most violations occur when the recovery policy is not activated, which indicates that the bottleneck in this task is the quality of the safety critic. In Navigation 2, \algname never violates constraints and only model-free recovery was run for \algname on the physical robot. In the Dynamic Obstacle Object Extraction environment, we observe a more even combination of low safety critic values and recovery errors during constraint violations.}
\label{table:violations}
\begin{center}
\resizebox{0.9\textwidth}{!}{%
\begin{tabular}{cccccccc}
\toprule
 & \textbf{Navigation 1} & \textbf{Navigation 2} & \textbf{Maze} & \textbf{Object Extraction} & \textbf{Object Extraction (Dynamic Obstacle)} & \textbf{Image Maze} & \textbf{Image Obstacle Avoidance}\\
\midrule
\textbf{MF Recovery }& N/A & N/A & $0.828\pm 0.115$ & $0.954 \pm 0.024$ & $0.550 \pm 0.049$ & $0.717 \pm 0.156$ & $0.000 \pm 0.000$\\ 
\textbf{MB Recovery} & N/A & $1.000 \pm 0.000$ & $0.858 \pm 0.039$ & $0.98344 \pm 0.01655$ & $0.269 \pm 0.055$ &  $0.583 \pm 0.059$ & N/A\\ 
\bottomrule
\end{tabular}}
\end{center}
\end{table*}

\section{Safety Critic Visualizations}
\label{sec:safety-critic-vis}
\begin{figure*}
     \centering
     \includegraphics[width=0.85\textwidth]{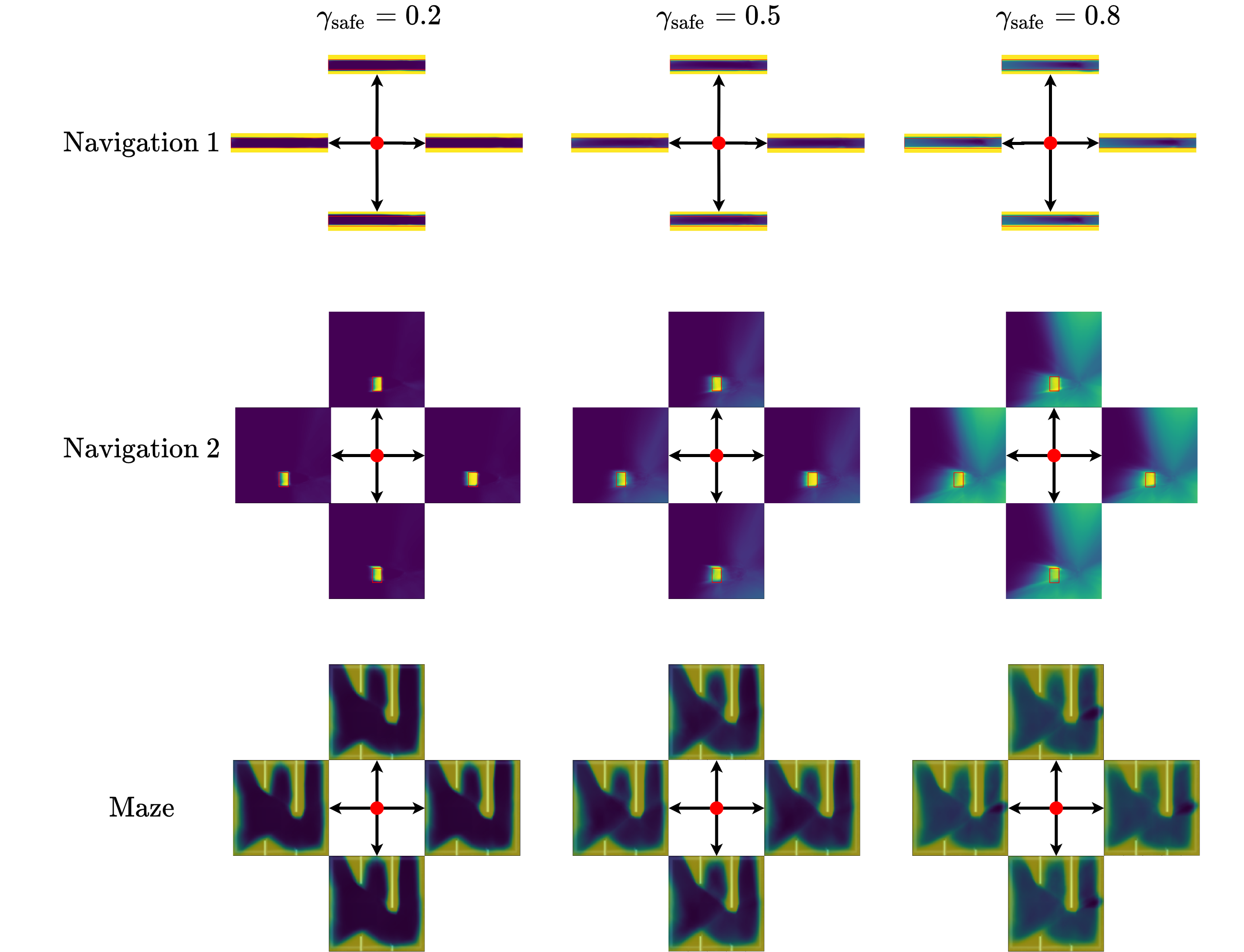}
     \caption{\textbf{$\qsafehat$ Visualization: }We plot the safety critic $\qsafe$ for the navigation environments using the cardinal directions (left, right, up, down) as action input. We see that as $\gsafe$ is increased, the gradient is lower, and the the function more gradually increases as it approaches the obstacles. Increasing $\gsafe$ essentially increases the amount of information preserved from possible future constraint violations, allowing them to be detected earlier. These plots also illustrate action conditioning of the safety critic values. For example, the down action marks states as more unsafe than the up action directly above walls and obstacles.}
        \label{fig:sim-exps-heatmaps}
\end{figure*}

We visualize the safety critic after pretraining for the navigation domains in Figure~\ref{fig:sim-exps-heatmaps} and observe that increasing $\gsafe$ results in a more gradual increase in regions near obstacles. Increasing $\gsafe$ carries more information about possible future violations in $\qsafe(s,a)$. However, increasing $\gsafe$ too much causes the safety critic to bleed too much throughout the state-action space as in the right-most column, making it difficult to distinguish between safe and unsafe states.

\section{Implementation Details}
\label{sec:imp-details}
Here we overview implementation and hyperparameter details for \algname and all comparisons. The recovery policy ($\pirec$) and task policy ($\pitask$) are instantiated and trained in both the offline phase, in which data from $\demos$ is used to pre-train the recovery policy, and the online phase, in which \algname updates the task policy with its exploration constrained by the learned safety critic and recovery policy. The safety critic and recovery policy are also updated online.

For all experiments, we build on the PyTorch implementation of Soft Actor Critic~\cite{SAC-applications} provided in~\cite{SAC_github} and all trained networks are optimized with the Adam optimizer with a learning rate of $3e-4$. We first overview the hyperparameters and training details shared across \algname and comparisons in Section~\ref{subsubsec:training-details} and then discuss the implementation of the recovery policy for \algname in Section~\ref{subsec:recovery-details}.

\subsection{Network Architectures}
\label{subsubsec:architecture}
For low dimensional experiments, we represent the critic with a fully connected neural network with 2 hidden layers of size 256 each with ReLU activations. The policy is also represented with a fully connected network with 2 hidden layers of size 256 each, uses ReLU activations, and outputs the parameters of a conditional Gaussian. We use a deterministic version of the same policy for the model-free recovery policy. For image-based experiments, we represent the critic with a convolutional neural network with 3 convolutional layers to embed the input image and 2 fully connected layers to embed the input action. Then, these embeddings are concatenated and fed through two more fully connected layers. All fully connected layers have 256 hidden units each. We utilize 3 convolutional layers, with 128, 64, and 16 filters respectively. All layers utilize a kernel size of 3, stride of 2, and padding of 1. ReLU activations are used between all layers, and batch normalization units are added for the convolutional layers. For all algorithms which utilize a safety critic (\algname, LR, SQRL, RSPO, RCPO), $\qsafe$ is represented with the same architecture as the task critic except that a sigmoid activation is added at the output head to ensure that outputs are on $[0, 1]$ in order to effectively learn the probability of constraint violation. The task and model-free recovery policies also use the same architectures for image-based experiments, except that they output the parameters of a conditional Gaussian over the action space or an action, respectively.

\subsection{Global Training Details}
\label{subsubsec:training-details}
To prevent overestimation bias, we train two copies of all critic networks to compute a pessimistic (min for task critic, max for safety critic) estimate of the Q-values. Each critic is associated with a target network, and Polyak averaging is used to smoothly anneal the parameters of the target network. We use a replay buffer of size $1000000$ and target smoothing coefficient $\tau = 0.005$ for all experiments except for the manipulation environments, in which a replay buffer of size $100000$ and target smoothing coefficient $\tau = 0.0002$. All networks are trained with batches of $256$ transitions. Finally, for SAC we utilize entropy regularization coefficient $\alpha = 0.2$ and do not update it online. We take a gradient step with batch size $1000$ to update the safety critic after each timestep. We also update the model free recovery policy if applicable with the same batch at each timestep. If using a model-based recovery policy, we update it for $5$ epochs at the end of each episode. For pretraining, we train the safety critic and model-free recovery policy for $10,000$ steps. We train the model-based recovery policy for $50$ epochs.

\subsection{Recovery Policy Training Details}
\label{subsec:recovery-details}
In this section, we describe the neural network architectures and training procedures used by the recovery policies for all tasks.
\subsubsection{Model-Free Recovery}
\label{subsubsec:model-free-recovery}
The model-free recovery policy uses the same architecture as the task policy for all tasks, as described in Section~\ref{subsubsec:architecture}. However, it directly outputs an action in the action space instead of a distribution over the action space and greedily minimizes $\qsafehat$ rather than including an entropy regularization term as in~\cite{SAC}. The recovery policy is trained at each timestep on a batch of $1000$ samples from the replay buffer.

\subsubsection{Model-Based Recovery Training Details}
\label{subsubsec:model-based-recovery}
For the non-image-based model-based recovery policy, we use PETS~\cite{handful-of-trials,PETS_github_pytorch}, which trains and plans over a probabilistic ensemble of neural networks. We use an ensemble of $5$ neural networks with 3 hidden layers of size 200 and swish activations (except at the output layer) to output the parameters of a conditional Gaussian distribution. We use the TS-$\infty$ trajectory sampling scheme from~\citet{handful-of-trials} and optimize the MPC optimization problem with $400$ samples, $40$ elites, and $5$ iterations for all environments. For image-based tasks, we utilize a VAE based latent dynamics model as in~\citet{nair2020goal}. We train the encoder, decoder, and dynamics model jointly where the encoder and decoder and convolutional neural networks and the forward dynamics model is a fully connected network. We follow the same architecture as in~\citet{nair2020goal}. For the encoder we utilize the following convolutional layers (channels, kernel size, stride): [(32, 4, 2), (32, 3, 1), (64, 4, 2), (64, 3, 1), (128, 4, 2), (128, 3, 1), (256, 4, 2), (256, 3, 1)] followed by fully connected layers of size $[1024, 512, 2L]$ where $L$ is the size of the latent space (predict mean and variance). All layers use ReLU activations except for the last layer. The decoder takes a sample from the latent space of dimension $L$ and then feeds this through fully connected layers $[128, 128, 128]$ which is followed by de-convolutional layers (channels, kernel size, stride): $[(128, 5, 2), (64, 5, 2), (32, 6, 2), (3, 6, 2)]$. All layers again use ReLU activations except for the last layer, which uses a Sigmoid activation. For the forward dynamics model, we use a fully connected network with layers $[128, 128, 128, L]$ with ReLU activations on all but the final layer.

\section{Environment Specific Algorithm Parameters}
\label{subsec:alg-params}
We use the same $\gsafe$ and $\esafe$ for LR, RSPO, SQRL, and RCPO. For LR, RSPO, and SQRL, we find that the initial choice of $\lambda$ strongly affects the overall performance of this algorithm and heavily tune this. We use the same values of $\lambda$ for LR and SQRL, and use twice the best value found for LR in as an initialization for the $\lambda$-schedule in RSPO. We also heavily tune $\lambda$ for RP and RCPO. These values are shown for each environment in Tables~\ref{table:hyperparams-outline} and~\ref{table:hyperparams}. 

\begin{table*}[t]
\caption{Hyperparameters for \algname and comparisons for all domains}
\label{table:hyperparams}
\begin{center}
\resizebox{0.8\textwidth}{!}{
\begin{tabular}{ c|c|c|c|c|c| } 
  & LR & RP & RCPO & MF Recovery & MB Recovery\\
 \hline
 Navigation 1 & $(0.8, 0.3, 5000)$ & $1000$ &  $(0.8, 0.3, 1000)$ & $(0.8, 0.3)$ & $(0.8, 0.3, 5)$\\ 
 Navigation 2 & $(0.65, 0.2, 1000)$ & $3000$ &  $(0.65, 0.2, 5000)$ & $(0.65, 0.2)$ & $(0.65, 0.2, 5)$\\
 Maze & $(0.5, 0.15, 100)$ & $50$ &  $(0.5, 0.15, 50)$ & $(0.5, 0.15)$ & $(0.5, 0.15, 15)$\\ 
 Object Extraction & $(0.75, 0.25, 50)$ & $50$ &  $(0.75, 0.25, 50)$ & $(0.75, 0.25)$ & $(0.85, 0.35, 15)$\\ 
 Object Extraction (Dyn. Obstacle) &$(0.85, 0.25, 20)$ & $25$ & $(0.85, 0.25, 10)$ & $(0.85, 0.35)$ & $(0.85, 0.25, 15)$\\
 Image Maze & $(0.65, 0.1, 10)$ & $20$ &  $(0.65, 0.1, 20)$ & $(0.65, 0,1)$ & $(0.6, 0.05, 10)$\\ 
 Image Obstacle Avoidance & $(0.55, 0.05, 1000)$ & N/A & N/A & $(0.55, 0.05)$ & N/A\\
\end{tabular}}
\end{center}
\end{table*}
\begin{table}
\caption{Hyperparameters for \algname and all comparisons.}
\label{table:hyperparams-outline}
\begin{center}
\resizebox{0.65\columnwidth}{!}{
\begin{tabular}{ c|c } 
 Algorithm Name & Hyperparameter Format\\
 \hline
 LR & $(\gsafe, \esafe, \lambda)$ \\ 
 RP & $\lambda$ \\ 
 RCPO & $(\gsafe, \esafe, \lambda)$ \\
 MF Recovery & $(\gsafe, \esafe)$\\
 MB Recovery & $(\gsafe, \esafe, H)$\\
 \hline
\end{tabular}}
\end{center}
\end{table}

\section{Environment Details}
\label{sec:exp-details}
In this section, we provide additional details about each of the environments used for evaluation.
\subsection{Navigation Environments}
\label{subsec:navigation-exp-details}
The Navigation 1 and 2 environments have linear Gaussian dynamics and are built from scratch while the Maze environment is built on the Maze environment from~\cite{nair2020goal}. In all navigation environments, offline data is collected by repeatedly initializing the pointmass agent randomly in the environment and executing controls to make it collide with the nearest obstacle.

\begin{enumerate}
    \item \textbf{Navigation 1 and 2:} This environment has single integrator dynamics with additive Gaussian noise sampled from $\mathcal{N}(0, \sigma^2 I_2)$ where $\sigma = 0.05$ and drag coefficient $0.2$. The start location is sampled from $\mathcal{N}\left((-50, 0)^\top, I_2\right)$ and the task is considered successfully completed if the agent gets within 1 unit of the origin. We use negative Euclidean distance from the goal as a reward function. Methods that use a safety critic are given $8000$ transitions of data for offline pretraining. For Navigation 1, $455$ of these transitions contain constraint violating states, while in Navigation 2, $778$ of these transitions contain constraint violating states.
    \item \textbf{Maze:} This environment is implemented in MuJoCo and we again use negative Euclidean distance from the goal as a reward function. Methods that use a safety critic are given $10,000$ transitions of data for offline pretraining, $1163$ of which contain constraint violating states.
\end{enumerate}
\subsection{Manipulation Environments}
    We build two manipulation environments on top of the cartgripper environment in the visual foresight repository~\cite{visualforesight}. The robot can translate in cardinal directions and open/close its grippers. In manipulation environments, offline data is collected by tuning a proportional controller to guide the robot end effector towards the objects. For the offline constraint violations, Gaussian noise is added to the controls when the end effector is sufficiently close to the objects to increase the likelihood of constraint violations. Additionally, to seed the task critic function for SAC to ease exploration for all algorithms, we utilize the same PID controller to collect task demos illustrating the red object being successfully lifted by automatically opening and closing the gripper when the end effector is sufficiently close to the red object.
\label{subsec:manip-exp-details}
\begin{enumerate}
    \item \textbf{Object Extraction:} This environment is implemented in MuJoCo, and the reward function is $-1$ until the object is grasped and lifted, at which point it is $0$ and the episode terminates. Constraint violations are determined by checking whether any object's orientation is rotated about the x or y axes by at least $15$ degrees. All methods that use a safety critic are given $20,000$ transitions of data for offline pretraining, $363$ of which contain constraint violating states. All methods are given $1000$ transitions of task demonstration data to pretrain the task policy's critic function. 
    \item \textbf{Object Extraction (Dynamic Obstacle):} This environment is implemented in MuJoCo, and the reward function is $-1$ until the object is grasped and lifted, at which point it is $0$ and the episode terminates. Constraint violations are determined by checking whether any object's orientation is rotated about the x or y axes by at least $15$ degrees. Additionally, there is a distractor arm that is moving back and forth in the workspace in a periodic fashion. Arm collisions are also considered constraint violations. All methods that use a safety critic are given $20,000$ transitions of data for offline pretraining, $896$ of which contain constraint violating states. All methods are given $1000$ transitions of task demonstration data to pretrain the task policy's critic function.
\end{enumerate}

\begin{figure}[htb!]
     \centering
     \includegraphics[width=0.43\textwidth]{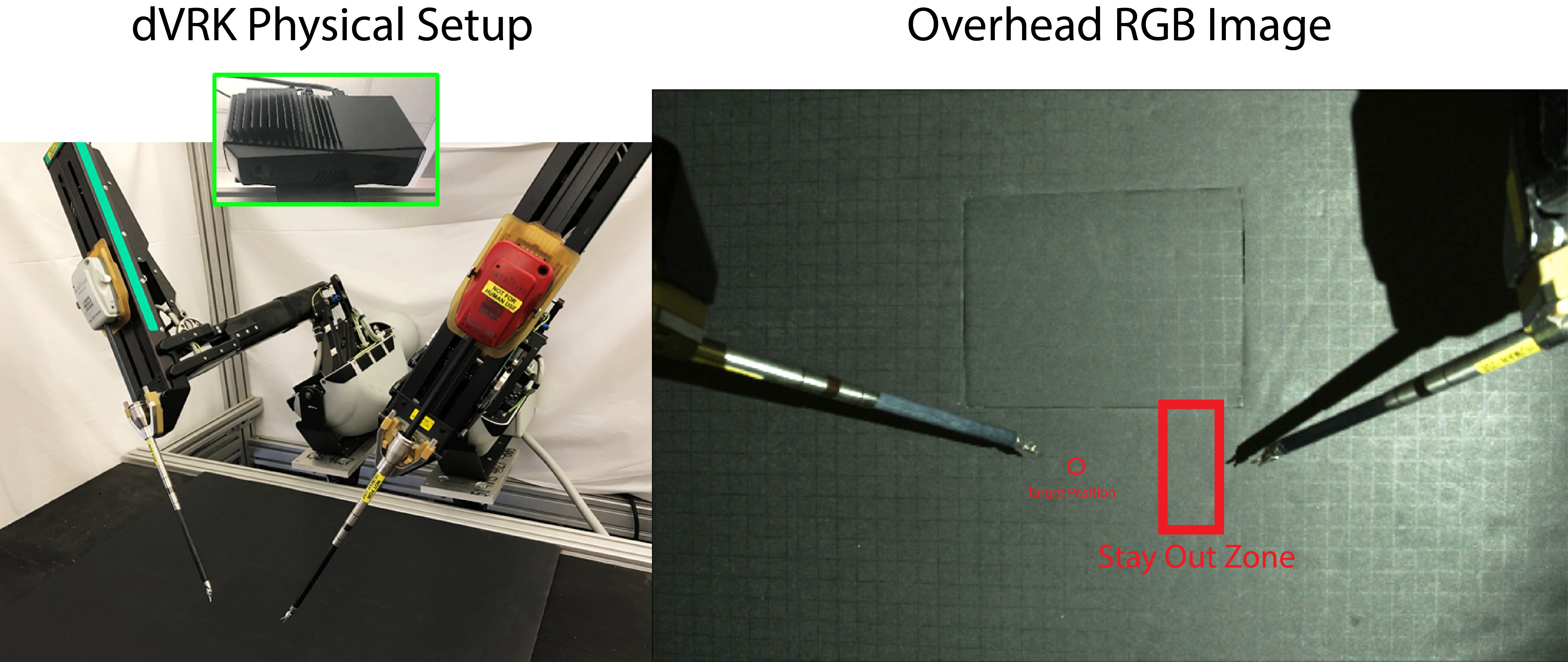}
     \includegraphics[width=0.43\textwidth]{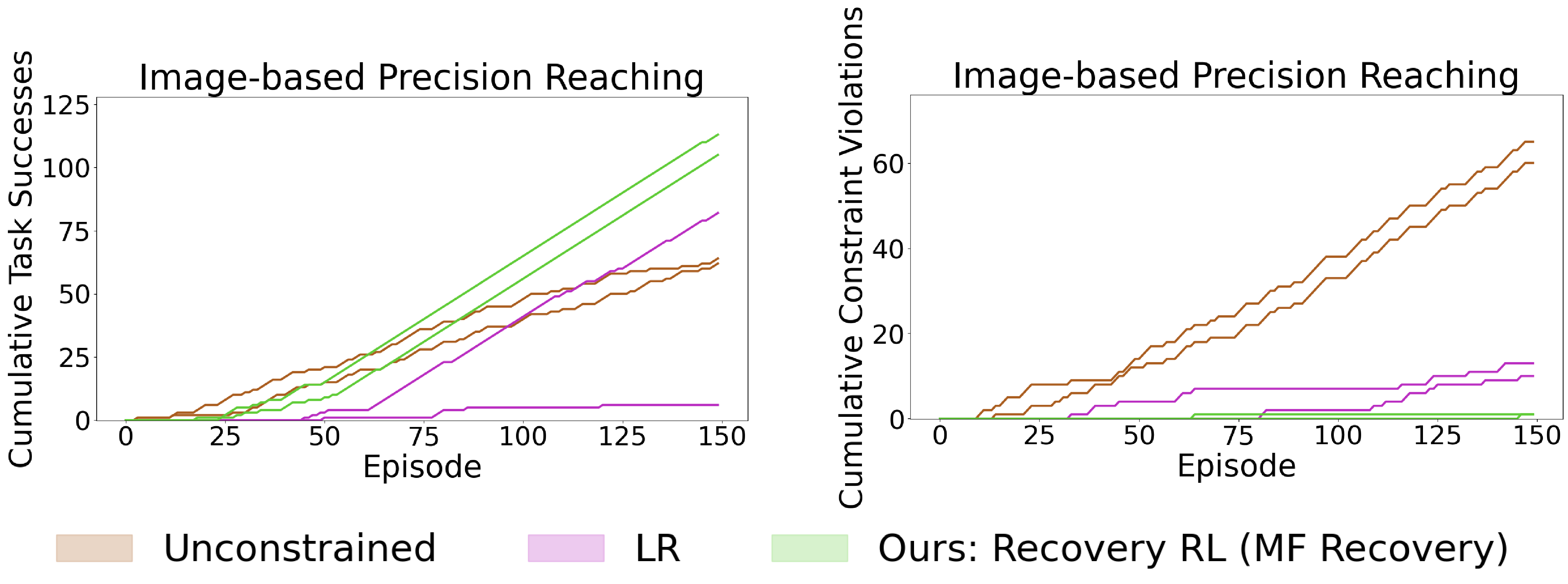}
     \caption{\textbf{Additional Physical Experiment: } The image reacher task on the dVRK involves guiding the end effector to a target position while avoiding an invisible stay out zone in the center of the workspace. We plot the cumulative constraint violations and task successes the image reacher task on the dVRK. We observe that \algname is both more successful and safer than LR and unconstrained.}
        \label{fig:phys-exps-old}
\end{figure}

\subsection{Image Maze}
\label{subsec:image-exp-details}
This maze is also implemented in MuJoCo with different walls from the maze that has ground-truth state. Constraint violations occur if the robot collides with a wall. All methods are only supplied with RGB images as input, and all methods that use the safety critic are supplied with $20,000$ transitions for pretraining, $3466$ of which contain constraint violating states.

\subsection{Physical Experiments}
\label{subsec:phys-exp-details}
Physical experiments are run on the da Vinci Research Kit (dVRK)~\cite{kazanzides-chen-etal-icra-2014}, a cable-driven bilateral surgical robot. Observations are recorded and supplied to the policies from a Zivid OnePlus RGBD camera. However, we only use RGB images, as the capture rate is much faster, and we subsample the images so input images have dimensions $48\times 64 \times 3$. End effector position is checked \textit{by the environment} using the robot's odometry to check task completion, but this is not supplied to any of the policies. In practice, the robot's end effector position can be slightly inaccurate due to cabling effects such as hysteresis~\cite{hwang2020efficiently}, but we ignore these effects in this paper. We train a neural network to classify whether the robot is in collision based on its current readings and joint position. All methods that use a safety critic are supplied with $6,000$ transitions of data for pretraining, $649$ of which contain constraint violating states. As for the navigation environments, offline data is collected by randomly initializing the end effector in the environment and guiding it towards the nearest obstacle. To reduce extrapolation errors during learning, we sample a start state on the right and left sides of the workspace with equal probability.

\subsection{Additional Physical Experiment}
We also evaluate \algname and comparisons on an image-based reaching task where the robot must make sure the end effector position does not intersect with a stay out zone in the center of the workspace instead of physical bumpers. The setup is almost identical to the setup described in Section~\ref{subsec:phys-exp-details}. We again provide RGB images to algorithms, and use 10,000 transitions to pre-train the safety critic. We again find that \algname is able to outperform comparisons on this task, both in terms of constraint satisfaction, and task completion.

\end{document}